\documentclass[sigconf,natbib=true]{acmart}
\usepackage{algorithm}
\usepackage{algorithmic}
\usepackage{multirow}
\usepackage{array}
\usepackage{float}
\usepackage{enumitem}
\usepackage{array}
\usepackage{makecell}
\usepackage{listings}
\newcolumntype{T}{>{\tiny}l} 
\newcolumntype{H}{>{\Huge}l} 
\newcolumntype{P}[1]{>{\centering\arraybackslash}p{#1}}

\AtBeginDocument{%
  \providecommand\BibTeX{{%
    \normalfont B\kern-0.5em{\scshape i\kern-0.25em b}\kern-0.8em\TeX}}}

\setcopyright{acmcopyright}
\copyrightyear{2023} 
\acmYear{2023} 
\setcopyright{acmlicensed}\acmConference[SIGIR-AP '23]{Annual International ACM SIGIR Conference on Research and Development in Information Retrieval in the Asia Pacific Region}{November 26--28, 2023}{Beijing, China}
\acmBooktitle{Annual International ACM SIGIR Conference on Research and Development in Information Retrieval in the Asia Pacific Region (SIGIR-AP '23), November 26--28, 2023, Beijing, China}
\acmPrice{15.00}
\acmDOI{10.1145/3624918.3625315}
\acmISBN{979-8-4007-0408-6/23/11}

\begin{document}

\title{Social Media Fashion Knowledge Extraction as Captioning}

\author{Yifei Yuan}
\affiliation{%
  \institution{The Chinese University of Hong Kong}
  \country{Hong Kong SAR}
}
\email{yfyuan@se.cuhk.edu.hk}

\author{Wenxuan Zhang}
\affiliation{%
  \institution{Alibaba DAMO Academy}
  \country{Singapore}
}
\email{isakzhang@gmail.com}

\author{Yang Deng}
\affiliation{%
  \institution{National University of Singapore	}
  \country{Singapore}
}
\email{dengyang17dydy@gmail.com}

\author{Wai Lam}
\affiliation{%
 \institution{The Chinese University of Hong Kong}
 \country{Hong Kong SAR}}
 \email{wlam@se.cuhk.edu.hk}

\renewcommand{\shortauthors}{}

\begin{abstract}
Social media plays a significant role in boosting the fashion industry, where a massive amount of fashion-related posts are generated every day. In order to obtain the rich fashion information from the posts, we study the task of social media fashion knowledge extraction. Fashion knowledge, which typically consists of the occasion, person attributes, and fashion item information, can be effectively represented as a set of tuples. Most previous studies on fashion knowledge extraction are based on the fashion product images without considering the rich text information in social media posts. Existing work on fashion knowledge extraction in social media is classification-based and requires to manually determine a set of fashion knowledge categories in advance. In our work, we propose to cast the task as a captioning problem to capture the interplay of the multimodal post information. Specifically, we transform the fashion knowledge tuples into a natural language caption with a sentence transformation method. Our framework then aims to generate the sentence-based fashion knowledge directly from the social media post. Inspired by the big success of pre-trained models, we build our model based on a multimodal pre-trained generative model and design several auxiliary tasks for enhancing the knowledge extraction. Since there is no existing dataset which can be directly borrowed to our task, we introduce a dataset consisting of social media posts with manual fashion knowledge annotation. Extensive experiments are conducted to demonstrate the effectiveness of our model. 
\end{abstract}





\maketitle

\section{Introduction}


Fashion, as one of the most important aspects of modern life, has been flourishing and evolving over the past decades. As a new style of sharing information, social media has become an important platform of constantly updating fashion information.
Given the rich fashion-related posts, extracting fashion knowledge from them can assist many downstream applications such as personalized recommendation~\cite{hu2015collaborative,kang2017visually,sigir21-crs,tkde-crs},  fashion image retrieval~\cite{liu2018deep,wang2018attentive,yuan2021conversational} and so on, therefore 
attracting increasing attention in recent years ~\cite{liu2018deep,wang2018attentive,yuan2021conversational}. 

\begin{figure}
    \centering
    \includegraphics[width=\linewidth]{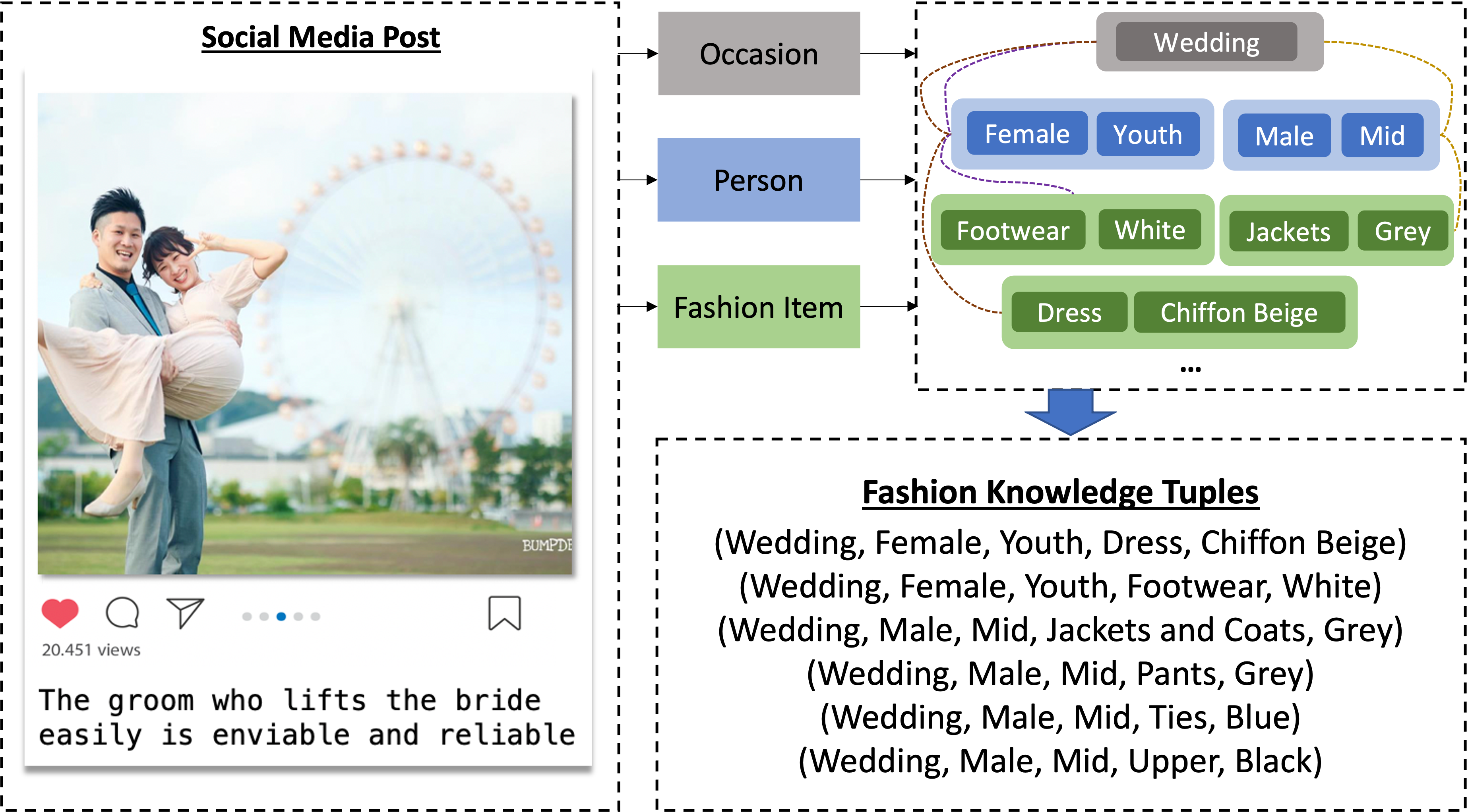}
    \caption{One example for the fashion knowledge extraction task. The fashion knowledge tuple consists of the occasion, person attributes including gender and age group, and the type and appearance of the fashion items in the post.}
    \label{fig1}
\end{figure}

As shown in Figure \ref{fig1}, a fashion post on the left side often consists of an image and the post text content written by the user for sharing his/her feelings. The Fashion Knowledge Extraction (FKE) task thus aims to elicit key fashion information from the post, including the occasion, person attributes, and the detailed fashion item information. Following the  previous study~\cite{ma2019and,re-survey}, the extracted fashion knowledge is usually denoted as a set of tuples containing essential fashion information. As listed on the right bottom part, the organized knowledge tuples  represent the fashion-related information from the post in a more structured manner.

Solving the FKE task on social media posts is an interesting yet challenging problem. Firstly, most existing FKE works are directly conducted on the fashion product images~\cite{jia2018deep,liu2016deepfashion,huang2015cross,chen2012describing} where a single image taken in the professional studio is provided. However, social media posts often contain information of different modalities, including both the image and text. As shown in Figure~\ref{fig1}, apart from the post image, the corresponding post text also indicates essential information such as where the image is taken, who in the post is, and what the person is wearing, and thus attaching great importance to extracting the fashion knowledge from the post. Therefore, how to make full use of the multimodal post information for the FKE task is  underexplored. 
To handle the multimodal information for harvesting the fashion knowledge, some initial attempts have been made. \citet{ma2019and} propose a pipeline-based model which first extracts person and clothing boxes from the image, then classifies the detected regions into different attribute categories with the text as an additional input. However, the model merely incorporates the image and text features by simple concatenation which fails to capture the deep interplay between different modalities. Moreover, similar to text-based structure prediction problems~\cite{acl21-absa,emnlp21-absa}, formulating the knowledge extraction task as a classification problem needs to manually determine a set of fashion knowledge categories in advance. However, the format of fashion knowledge aspects is typically quite varied. For example, ``muslin white'' and ``chiffon beige'' can both be used to describe the appearance of the dress the woman wears in Figure \ref{fig1}. Besides, strong dependencies are often  observed between different aspects in the fashion knowledge data~\cite{ma2019and}. For instance, the person clothes can be affected by the occasion of the post. Traditional classification-based models tend to determine the category of each fashion knowledge aspect separately, thus failing to capture such relationship. Furthermore, their method is pipeline-based and can give rise to the problem of error propagation. Some potential errors in the preceding steps  such as the inaccurate prediction of person boxes could lead to a negative influence on not only the extraction of person attributes information but also all the fashion item knowledge corresponding to the person. 



To tackle the research challenges discussed above, we propose to cast the social media based FKE task as a captioning problem. Inspired by the classic image captioning task~\cite{vinyals2015show,xu2015show} that generates a natural language description for a given image, we transform the FKE task to a captioning problem for better modeling the interplay of the image and text information and alleviating the issues of the classification-based models. 
Specifically, given the multimodal social media post including an image and the corresponding text, we aim to generate a natural language caption for the post, which contains the key fashion information. The fashion knowledge tuples can then be easily extracted from the generated caption. During the training stage, we first transform the original fashion knowledge tuples into a pseudo caption with a sentence transformation method. Then the multimodal post and the pseudo caption can be paired as training instances to learn a multimodal generation model.
With such caption generation formulation, we can tackle the FKE task in an end-to-end manner, alleviating the potential error propagation issue in pipeline-based  solutions.
Moreover, compared with existing classification-based models, our model incorporates the multimodal information from both the image and text as input and utilizes the natural language caption as the output, which can better capture the interactions between different modalities. In addition, the dependencies between different fashion knowledge aspects can also be fully exploited by learning to generate them in an autoregressive manner. 

Motivated by the big success of  pre-trained language models for various vision-language tasks such as image-text retrieval~\cite{lu2019vilbert}, we build our model based on a multimodal pre-trained generation model named  VL-Bart~\cite{cho2021vlt5}  to utilize its rich knowledge of processing information from different modalities. 
We further design several auxiliary tasks including visual question answering (VQA), sentence reconstruction, and image-text matching to warm-up the model. These tasks are designed to equip the model with fashion-related knowledge via different formats but under the same model architecture. After training with multiple relevant tasks, the model can obtain some prior task-specific knowledge, which helps tackle the main concerned FKE task.

Since existing datasets used in previous studies are either single-modal with only fashion item information  \cite{liu2016deepfashion} or not publicly available \cite{ma2019and}, there is no dataset that can be directly adopted for the concerned task.
Therefore, we introduce a large-scale fashion knowledge dataset based on user-generated social media fashion-related posts. For each post including an image and text, we manually annotate its corresponding occasion, person attributes, as well as the type and appearance of the fashion items they wear to construct the fashion knowledge tuples. We provide detailed statistics on this newly introduced dataset and conduct extensive experiments on it\footnote{The dataset and code are available in https://github.com/yfyuan01/FKE.}.

To sum up, the main contributions of our paper are as follows:
\begin{itemize}
\item We propose to tackle fashion knowledge extraction from multimodal social media posts as a captioning task, which effectively captures the interplay of different modalities via generating a natural language caption for extracting the fashion knowledge tuples in an end-to-end manner.
\item To equip the model with  fashion-related knowledge, we design several auxiliary tasks  including sentence reconstruction, image-text matching, and visual question answering, which helps  
tackle the main concerned FKE task.
\item We contribute a benchmark dataset and conduct extensive experiments to demonstrate the effectiveness of our model. We show that our method outperforms various state-of-the-art methods, especially under the difficult multi-person multi-fashion-item situation.
\end{itemize}

\section{Related Work}
\subsection{Fashion Knowledge Extraction}
Fashion knowledge plays a vital role in fashion-related tasks such as clothing recognition~\cite{chen2012describing,kalantidis2013getting}, fashion trend forecasting~\cite{ma2020knowledge,mall2019geostyle,matzen2017streetstyle}, fashion sentiment analysis~\cite{Yuan2021SentimentAO}, and fashion-related information retrieval~\cite{yuan2021conversational,wu2021fashion}. Therefore, there has been an increasing interest on knowledge extraction tasks in the fashion domain recently~\cite{huang2015cross,jia2018deep}. Early studies mostly rely on handcrafted features and mainly focus on extracting simple clothing-related knowledge using techniques such as conditional random field ~\cite{chen2012describing,liu2016deepfashion}. ~\citet{huang2015cross} propose a Dual Attribute-aware Ranking Network (DARN) consisting of two sub-networks for retrieval feature learning. DeepFashion, which is first proposed by ~\citet{liu2016deepfashion}, is annotated with clothing items with rich fashion knowledge information. They propose a dataset where each picture is annotated with some fashion item attributes. ~\citet{jia2018deep} propose a data-driven approach for recognizing fashion attribute where a modified version of Faster R-CNN model is trained. Furthermore,  ~\citet{wang2018attentive} solve the problem of fashion landmark localization and clothing category classification via a knowledge-guided fashion network. ~\citet{yan2017unconstrained} address unconstrained fashion landmark detection, where clothing bounding boxes are not provided in both training and testing phases. To the best of our knowledge, ~\citet{ma2019and} are the first to focus on social media based fashion knowledge extraction, which aims to conduct automatic fashion knowledge extraction from social media posts by unifying the occasion, person attributes and clothing prediction in a contextualized module. Although the model incorporates multimodal information from the social media posts, it is pipeline-based and requires to extract all person boxes in advance. Moreover, they do not publish their dataset for safety reasons.

\begin{figure*}
    \centering
    \includegraphics[height=55mm]{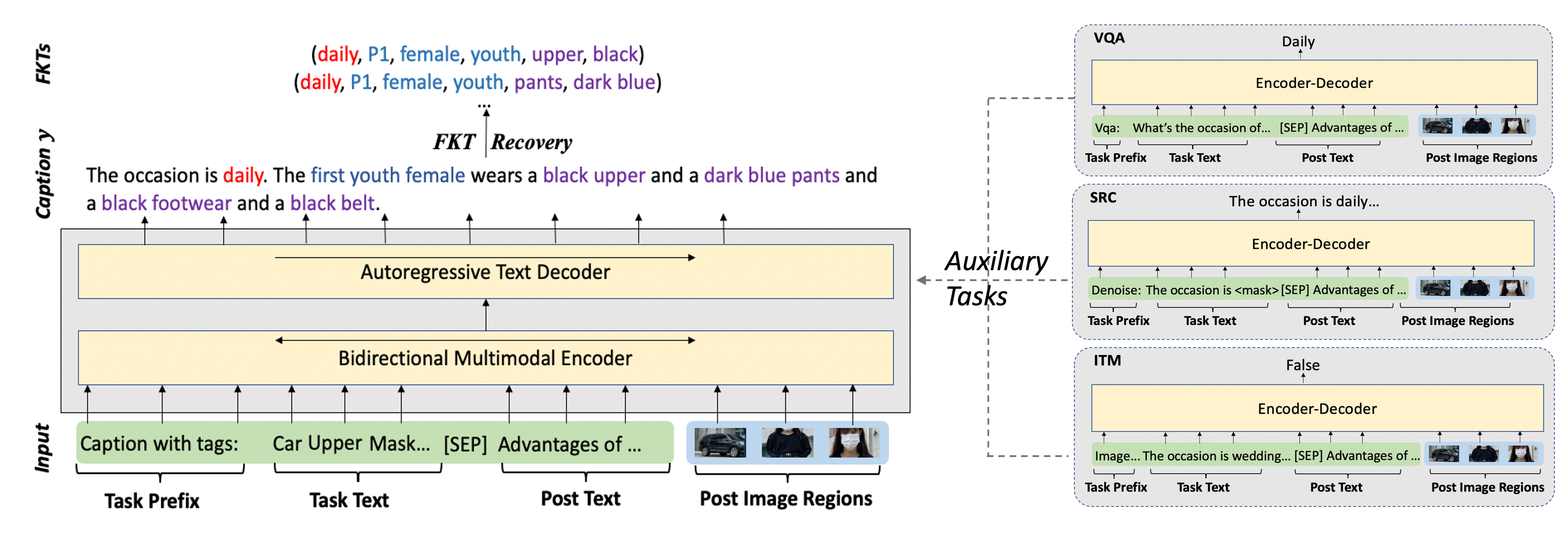}
    \caption{The overall architecture of our framework. The left part depicts transforming the FKE task as a captioning problem, where FKT denotes the fashion knowledge tuples. The right part shows the detail of the  three auxiliary tasks.}
    \label{fig:2}
\end{figure*}

\subsection{Multimodal Pre-training}
Following the success of large pre-trained  models in natural language understanding (NLU) ~\cite{devlin2018bert,yang2019xlnet,clark2020electra} and generation (NLG) tasks  ~\cite{lewis2019bart,raffel2019exploring,brown2020language}, some multimodal pre-trained models have shown their superiority over traditional non-pretrained methods in many  tasks recently. Some of them mainly focus on video-text pretraining such as  VideoBERT~\cite{DBLP:conf/iccv/SunMV0S19}, HERO ~\cite{li2020hero}, MIL-NCE ~\cite{miech19endtoend,miech19howto100m} and so on, while others focus on the image-text domain. Among these image-text pretrained methods, ViLBERT ~\cite{lu2019vilbert}, LXMERT ~\cite{yang2019xlnet}, and VL-BERT ~\cite{Su2020VL-BERT:} are the extensions of the popular BERT model~\cite{devlin2018bert} and are used for learning task-agnostic joint representations of the image content and natural language. Following this line, unified models are proposed to deal with both understanding and generation tasks. For example, Oscar ~\cite{li2020oscar} leverages object tags detected in images as anchor points to significantly ease the learning of image-text alignments. CLIP ~\cite{unpublished2021clip} connects image and text representations by learning visual concepts from natural language supervision. ~\citet{huo2021wenlan} propose a two-tower pre-trained model named WenLan within the cross-modal contrastive learning framework. CogView ~\cite{ding2021cogview} and DALLE~\cite{ramesh2021zeroshot} are powerful generative models that focus on text-to-image generation. Among them, VL-Bart ~\cite{cho2021vlt5} is the state-of-the-art model designed for vision text generation and shows good generalization ability on different tasks. Therefore, we adopt it as the backbone of our model (to leverage its knowledge in processing information from different modalities) in this work.

\section{Our Method}

\subsection{Problem Definition}
We aim to automatically extract fashion knowledge from a social media post, which is composed of an image and the post text content. Following the definition given in previous study~\cite{ma2019and}, the fashion knowledge is denoted as a set of tuples, each tuple $k$ is defined as the combination of the occasion, person attributes, and fashion item information: $k = (o, p, f)$. Here $o$ denotes the occasion category, which belongs to a set of occasions such as wedding, school, sports, etc. $p=(age, gender)$ denotes the gender and age information of a specific person in the post, where $gender\in\{Male, Female\}$ and $age\in\{Kid, Youth, Mid, Old\}$. The fashion item information $f=(type, app)$ contains the fashion item type $type$ such as ``pants'' and the appearance $app$ of the fashion item, where the appearance is usually a short text such as ``lace white''  describing both its pattern, color, and style. Therefore, the fashion knowledge tuple $k$ can also be unfolded and represented as $k=(occ,age,gender,type,app)$.

Given a post $x$ consisting of an image $v$ and text $t$  denoted as $x = \{v, t\}$, the problem is to develop a framework which outputs $N$ fashion knowledge tuples of the post, represented as $K=\{k\}_{i=1}^N$, where the number of tuples $N$ varies from post to post. 

\subsection{Framework Overview} 
Figure \ref{fig:2} presents an overview of our proposed method. In general, we formulate the concerned FKE task as a captioning problem. We tackle it via an encoder-decoder structure based on a pre-trained generative model named VL-Bart~\cite{cho2021vlt5}, as shown in the left part. By treating it as a multimodal generation problem, the interactions between different modalities can be effectively captured. Then the structured fashion knowledge tuples are recovered from the generated caption. Besides, as shown in the right part, before captioning, we leverage several fashion-related auxiliary tasks to warm-up the pre-trained models and equip it with task-specific knowledge.

In detail, for the captioning phase in the left   part, we fine-tune the model to generate the fashion knowledge captions. Instead of generating the tuple-like fashion knowledge directly, the model generates captions in a natural language manner. To facilitate such training, given the original training instance with the format of post-tuple pair $(x,K)$, we transform the fashion knowledge tuples $K$ to a pseudo caption $y$ containing all the desired fashion knowledge elements of the post via a caption construction method. The transformed training instance can thus be represented as $(x,y)$ for learning a multimodal generative model. To add fashion information to the pre-trained model, as shown in the right part of Figure \ref{fig:2}, we design various auxiliary tasks processes including  sentence reconstruction (SRC), visual question answering (VQA), and image text matching (ITM) before fine-tuning. These tasks are designed to focus on one or several fashion knowledge aspects and equip the model with  task-specific fashion knowledge.

\subsection{Image and Text Encoding}

\subsubsection{Text Encoding}
As shown in the bottom part of Figure \ref{fig:2}, the input text $t$ of our model consists of three parts: task prefix, task text, and post text. Post text is the original text content written by the user in the social media post. To   auxiliary task training, we also include a task prefix which indicates which task the model should perform, followed by  the task text used as an additional textual input for a specific task (e.g. it can be a question in the visual question answering task). The three textual inputs are concatenated with a special token [SEP] and fed to the embedding layer to obtain the text embedding of the model. The positional embeddings for denoting the absolute token positions are added to the token embeddings and learned during the training. Then for each training instance, the text input $t$ is encoded to a vector  represented as $e^t$.

\subsubsection{Image Encoding} 
\label{imageembedding}
To extract image features, we first detect several object regions from the image, denoted as Region of Interest (ROI). By utilizing ROI instead of the raw image pixels, we can align the multimodal information between the image and text~\cite{lu2019vilbert,li2020oscar}.
To obtain ROI features, following previous studies~\cite{lu2019vilbert,li2020oscar}, we generate $r$ image object regions with Faster-RCNN~\cite{renNIPS15fasterrcnn}. 
For each region, we also detect the object tag in the format of text such as ``Upper'', ``Woman'', etc. The final embedding is the sum of four types of features: ROI object features, ROI bounding box coordinates, image ids, and region ids. The ROI object feature is the encoding result from Faster-RCNN. The bounding box coordinate is the position vector of the ROI. Image id is set to be 1 in our task, and region id $\in\{1,...,r\}$. The visual embedding of image $v$ is represented as $e^v$.
%


\subsection{FKE as Captioning}  \label{pm}
We cast the original FKE task as a captioning problem. We aim to train a generation model for learning the mapping function given the natural language caption $y$ transformed from the fashion knowledge tuples $K$ and the social media post $x$.
\subsubsection{Caption Construction}
To facilitate the training process of the generative model, we propose a strategy to construct the pseudo caption $y$ from the $N$ fashion knowledge tuples of a post represented as $K = \{k\}_{i=1}^N$.
For the caption construction, we wish to incorporate the major fashion knowledge elements into the caption while neglecting the unnecessary information. The rule of transforming the fashion knowledge tuples into the natural language caption is  designed as  follows. 

As shown in Algorithm \ref{alg1}, since the occasion of all the fashion knowledge tuples corresponding to a certain post is the same, we first transform the occasion information into a sentence at the beginning of the target  sequence with the template ``\path|The|  \path|occasion| \path|is|  \verb|[occ]|''. In the example shown in Figure \ref{fig:2}, the sentence saying ``The occasion is daily'' is constructed to incorporate the occasion category. We then group and gather all the fashion knowledge tuples by different persons. For each person, we write a sentence containing his/her gender and age information. With the same example, we write ``The first youth female'' at the beginning of the second sentence. We then list all the fashion items the person wears including their  type and appearance and incorporate them into a fashion item description sentence. For different fashion items of the same person, we concatenate them with the word ``\verb|and|'' to mimic the writing method the users often use. Therefore, for the girl in the Figure \ref{fig:2}, we add the fashion item information by saying that she wears a black upper and a dark blue pants, etc.
After the sentence transformation process that transforms the original tuple-like data into a natural language caption, the input-to-target generation can be modeled with a classic encoder-decoder architecture. 

\begin{algorithm}[t]
\caption{Caption Construction} 
\label{alg1} 
\begin{algorithmic}[1] 
\small
\REQUIRE~~\\
$N$ fashion knowledge tuples of a post $\{k\}_{i=1}^N$\\
Each tuple $k=(occ, gender, age, type, app)$\\
Number of persons in the post $n_p$
\ENSURE~~\\
Natural Language Sequence $y$\\
\STATE $o\gets$ ``The occasion is ''$+occ$ \\
\STATE $s\gets o$
\FOR{$m=1$ to $n_p$}
\STATE $y \gets y$+ ``The '' +$m$+$gender$+$age$+``person wears ''
\FOR{$n=1$ to $N$}
\IF{$n!=N$}
\STATE$y \gets y$+ ``a'' + $app$ + $type$ + `` and ''
\ELSE
\STATE$y \gets y$ ``a'' + $app$ + $type$ + ``.''
\ENDIF
\ENDFOR
\ENDFOR
\end{algorithmic}
\end{algorithm}



\subsubsection{Encoder-Decoder Structure}
We use transformer ~\cite{vaswani2017attention} encoder-decoder to incorporate image and text features and generate the fashion knowledge caption. The encoder is composed of $m$ transformer blocks, each of which consists of a self-attention layer and a fully-connected layer with residual connections. The decoder is also a stack of $m$ transformers with an additional cross-attention layer in each block. Given the multimodal post input $x$, the image $v$ and text $t$ are first fed into the bidirectional encoder and incorporated together into a contextualized sequence. Given the sequence, the decoder models the conditional probability distribution of the target sentence to generate caption $y$. At each time step, the decoder iteratively predicts the probability of current caption tokens based on previously generated tokens and the encoder output.

\subsubsection{Training} 

\label{training}
Given a pretrained model with the encoder-decoder structure, we fine-tune our model parameters $\theta$ on the input-target pair. We utilize standard sentence generation loss as our loss function. At each time step $j$, the decoder output $y_j$ is determined based on the generated caption by previous time steps $y_{<j}$, the image and text embedding $e^v$ and $e^t$. We minimize the negative log-likelihood of generating the target caption $y$ given the input text embedding $e^t$ and image embedding $e^v$:
\begin{equation}
    {\rm min}\;-logP_\theta(y|e^t,e^v) = -\sum^{|y|}_{j=1}logP_\theta(y_j|y_{<j},e^t,e^v)
\end{equation}
where $P_\theta$ is the likelihood of generating the target caption $y$ given the image text input, and $|y|$ is the length of the target caption.

\subsubsection{Inference and Tuple Recovery}
During inference, we generate the target caption sequence $y'$ in an autoregressive manner given the post image and text pair. Same as mentioned in the training phase, the input text also consists of the task and the post text separated by the separation token [SEP].

At each time step, we choose the token with the highest probability over the vocabulary set to obtain the natural language caption. When recovering the fashion knowledge tuples  from the caption, we first split the output sequence into several sentences. As shown in the top left part of Figure \ref{fig:2}, the occasion information can then be extracted from the first sentence having the format of  ``\verb|The occasion is |''. With respect to the remaining sentences, we extract the person attributes in the sentence, and pair them with all the fashion item information including the type and appearance in that sentence. According to the figure, for the sentence   ``\verb|The first youth female wears a black upper|'', we can obtain the fashion knowledge tuple \verb|(youth,female,upper,black)| from it following the rules. After extracting the fashion knowledge elements from the sequence, we compare them with the ground-truth label for evaluation. Notably, if the decoding fails, say the generated sequence violates the format, we treat the prediction as null.

\subsection{Auxiliary Task Training}
\begin{figure}
    \centering
    \includegraphics[width=\linewidth]{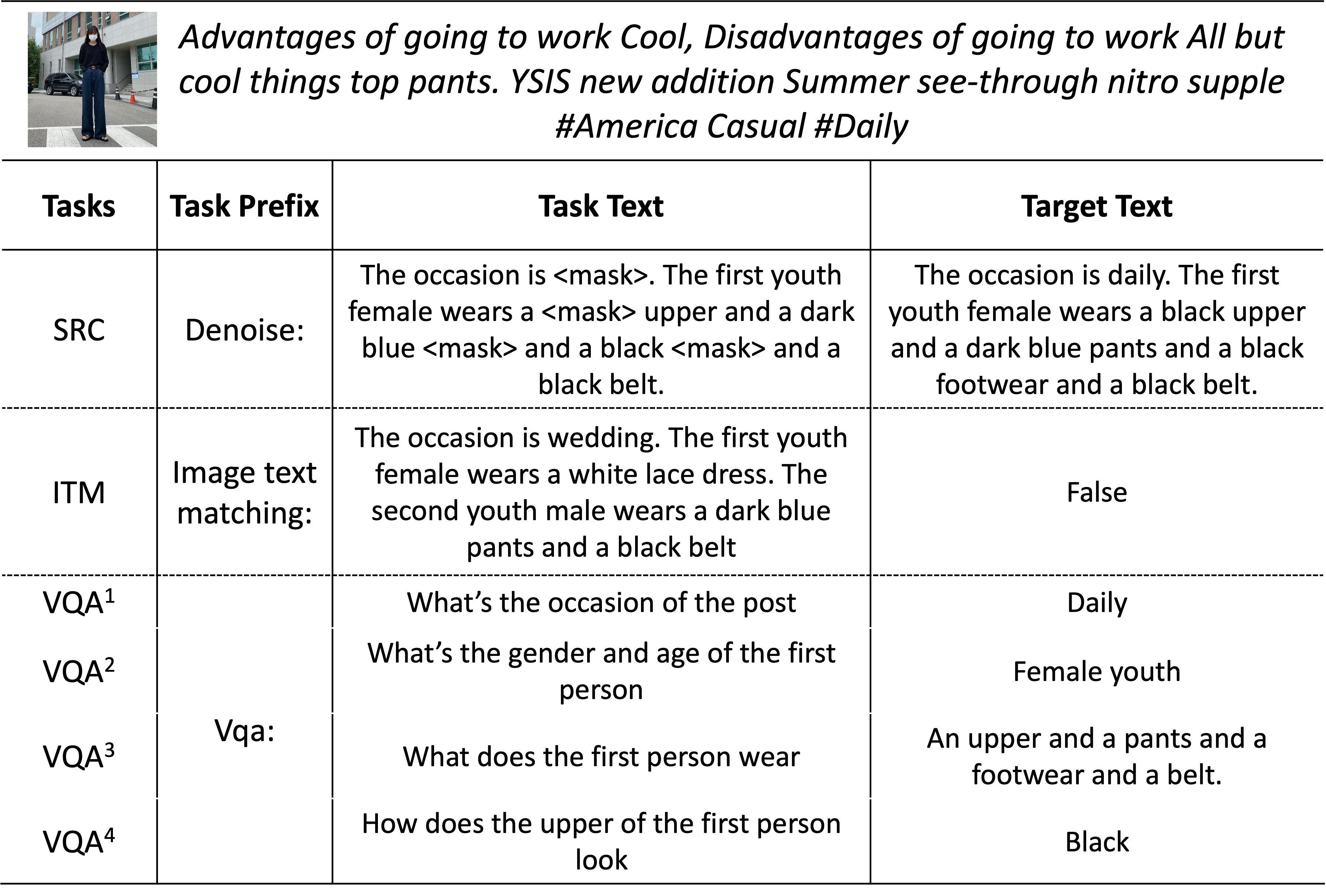}
    \caption{The task prefix and Input-Output formats of our three auxiliary tasks.}
    \label{fig:posttraining}
\end{figure}

To obtain task-specific knowledge, we further design several auxiliary tasks including sentence reconstruction, visual question answering, and image-text matching. These tasks are designed to focus on one or several fashion knowledge aspects which can  warm up the pre-trained model before training on the main captioning task. In order to fit to different auxiliary tasks, we assign different task prefixes to each task and add them before the original task text. The examples of the task prefix, task text, and target output text of each task are listed in Figure \ref{fig:posttraining}. 

\subsubsection{Sentence Reconstruction (SRC)} Based on the assumption that different aspects in the fashion knowledge data are not strictly independent but strongly related (e.g. the type and appearance of the fashion item can be affected by the occasion), the goal of the SRC task is to predict some masked tokens based on their surrounding tokens and the image feature. Therefore, we randomly mask out 30\% of the input tokens and ask the model to predict and reconstruct the original sentence. The task text is the masked fashion knowledge sequence and the output is the original full text. For each masked token, we replace it with the special mask token <mask>. 

\subsubsection{Image-Text Matching (ITM)} This task takes a pair of image and natural language text as input. The model needs to determine if the text corresponds to the image or not. In our setting, we aim to determine if the given fashion knowledge caption corresponds to the post or not. We transform the original binary classification task into a generation problem following the rule that if the text is the corresponding caption of the post, the model generates ``true'', while if not, the model generates ``false''. We consider the ground-truth post-caption pair as positive samples. To construct negative samples, with the probability of 50\%, we randomly sample the pseudo caption from another post in the training dataset.
\begin{figure*}
    \centering
    \includegraphics[width=0.8\linewidth]{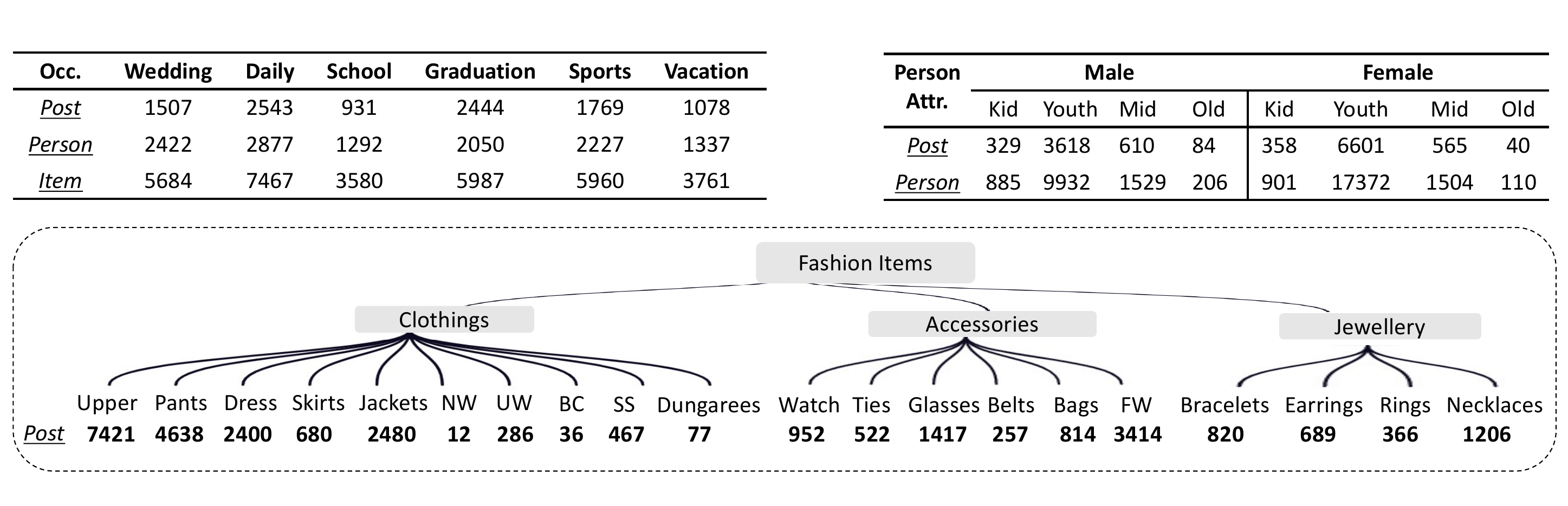}
    \caption{Detailed information of our dataset w.r.t different fashion knowledge aspects, where NW, UW, BC, SS, FW are the abbreviations of nightwear, underwear, babyclothes, swimsuits, and footwear respectively. From bottom to top, we report the number of fashion items, persons, and posts.}
    \label{fig:data}
\end{figure*}
\subsubsection{Visual Question Answering (VQA)} 
In the general visual question answering problem~\cite{antol2015vqa}, the model aims to generate the answer of the input question for a given image. In our setting, we design four types of question-answering pairs which correspond to the occasion, person attributes, and fashion item aspects of each post. Each data sample in the training set has one-quarter probability of transforming into one of the four QA pairs during training. As listed in Figure \ref{fig:posttraining}, the first type of question has a fixed form, which says ``what's the occasion of the post'', and the answer is constructed based on the occasion category in the ground-truth training dataset. The second type of question focuses on the person information of the post, which asks the gender and age group of a random person in the post. For example, ``what's the gender and age of the first person''. The third and fourth types of questions are related to the type and appearance of a random fashion item in the post. With respect to the third type of question, we first randomly select a person in the post, then asks what he/she is wearing. Considering the appearance of the fashion items, with the random selection of one fashion item in the post, we set the format of the questions as ``how does the [Fashion Item Type] of the [Person ID] person look''.


\subsubsection{Multitask Training} 
For multitask training, we train the model on different auxiliary tasks with the pretrained parameter weights. For each training step, we randomly sample a mini-batch from one of the three tasks. We differentiate the tasks by using different task prefixes. It is worth noting that the four subtasks of VQA are considered as the same task and share the same task prefix. Since we only change the input-output format without changing the pretrained model structure, we use the same loss function as in Section \ref{training}. We then set a bunch of weights according to the partial loss of each task to form the final loss. 
\begin{equation}
    L_{all} = \sum_{i=1}^{|T|}w_iL_i
\end{equation}
Where $|T|=3$ is the number of tasks, $L_i$ is the partial loss according to each task, $w_i$ is the hyperparameter representing the corresponding weight. After using  with multiple relevant tasks to warm-up the model, the model is equipped with fashion-related knowledge, which can help tackle the main concerned FKE task.


\section{Experiment}
\subsection{Dataset}
Since there is no existing dataset that can be directly adopted to our setting, we collect and contribute a large-scale annotated dataset for the FKE task. Our dataset contains 9,272 posts with 32,439 fashion knowledge tuples in total, with an  average of 3.5 fashion knowledge tuples per post and 2.7 fashion knowledge tuples per person. The detailed statistics of our dataset concerning different fashion knowledge aspects are reported in Figure~\ref{fig:data}.

\noindent\textbf{Post Collection and Preprocessing} Our dataset is collected from Instagram\footnote{\url{https://www.instagram.com/}}, which is a popular social media platform where large amount of posts are generated by users every day. To obtain the fashion-related posts, we first define six occasions, including school, graduation, sports, wedding, daily wear, and vacation. Under each occasion, we then choose some typical hashtags and crawl the related posts given the hashtag. After that, we filter out posts without any texts or containing only emojis.
Since the raw text in social media is often noisy, we employ several text cleaning methods to deal with the crawled texts. We first preprocess the texts by removing all the unnecessary tokens including emojis, URL, whitespace, HTML characters, punctuation marks, and mentions. We then detect and translate all the text into English using the Google translate API \footnote{https://pypi.org/project/googletrans/}.   


\noindent\textbf{Fashion Knowledge Annotation} For the filtered fashion-related posts, we hire 10 fashion experts to manually annotate the fashion knowledge information for each post. The annotators first need to determine the occasion of the posts. Since sometimes similar images may result in different occasion results, before making the choice, the annotators are asked to read both texts and images to make sure that the occasion type is determined by both of them. After that, the annotators annotate the person attributes including the gender and age group in the images, as well as the type and appearance of the fashion items they wear.  Considering the unfixed format of the appearance, when annotating it, the annotators are asked to use two or three words to describe how the fashion item looks, including its color, pattern, and texture. After all the annotations are finished, we ask two annotators to check the completeness and correctness of the results, making sure that all the fashion knowledge is correctly annotated in each post.


\begin{table*}[h!]
\caption{The experimental results of our model compared with the baseline methods, as well as the ablated results where text and image tags are removed from our model.}
\small
\begin{tabular}{p{2.4cm}<{\centering}|p{0.55cm}<{\centering}p{0.55cm}<{\centering}p{0.55cm}<{\centering}p{0.55cm}<{\centering}|p{0.55cm}<{\centering}p{0.55cm}<{\centering}p{0.55cm}<{\centering}p{0.55cm}<{\centering}|p{0.55cm}<{\centering}p{0.55cm}<{\centering}p{0.55cm}<{\centering}p{0.55cm}<{\centering}|p{0.55cm}<{\centering}p{0.55cm}<{\centering}p{0.55cm}<{\centering}p{0.55cm}<{\centering}}
\hline                                                  \multirow{2}{*}{Model} & \multicolumn{4}{c|}{\textbf{Occasion}} & \multicolumn{4}{c|}{\textbf{Category}} &
\multicolumn{4}{c|}{\textbf{Appearance}} &
\multicolumn{4}{c}{\textbf{Overall}} \\
    & Acc & Pre         & Rec           &F1 
    & Acc   & Pre     & Rec  &F1      
    & Acc   &Pre    & Rec  &F1
    & Acc   &Pre    & Rec  &F1\\ 
 \hline
    
    DARN ~\cite{huang2015cross}    & 44.1  & 40.2   & 42.6  & 42.4 & 25.1 & 73.2& 47.1 & 57.4& 10.3& 64.2 & 40.8 &50.0&9.1& 23.5& 14.4& 17.9\\
    FashionNet~\cite{liu2016deepfashion}     & 43.2 & 40.5&43.6 & 42.0 &26.3 &72.8&46.3&56.6&10.6 & 62.9 & 41.2 & 49.8 &8.7& 22.4 & 14.5& 17.6\\
    HDF ~\cite{ma2019and}    &50.3  & 47.4 & 43.7& 45.5 & 29.4 &77.1& 52.7 & 62.8&14.3&
    68.8&44.0&53.7& 12.1 & 27.9 &17.6 &21.6\\
    ViLBERT~\cite{lu2019vilbert}   & 59.6 & 50.3 & 58.4 & 54.1 &  32.5   & 80.1 &53.6 &64.2 &15.2 & 71.3 & 52.9 & 60.7 & 12.5 &28.7 & 20.4&23.5\\
    Oscar~\cite{li2020oscar} & 75.6 & 75.2 & 76.0 & 75.5 & 7.7 & 21.1& 33.7 & 26.0& 7.1 & 20.1 & 23.2 &21.5 & 5.5 & 15.2& 17.4& 16.2\\
    VL-Bart~\cite{cho2021vlt5}  & 75.2 & 69.1 & 74.9 & 71.4&30.8 & 80.9 &48.6 & 60.7 &17.8 & 52.9 &31.6&39.6&15.4&35.6&21.9& 27.1\\
\hline
    Ours w/o text & 69.2 & 64.6 & 70.1 & 68.8 & 32.7 &78.5 &64.2 & 70.6& 20.4&71.8& 57.7& 64.0& 15.4&33.6&28.2&30.7\\
    Ours w/o img tags & 72.8 & 68.4 & 73.0 & 70.6 & 30.8 &75.6 &64.2 & 69.4& 18.1&70.2& 57.0& 62.9& 16.0&35.1&28.4&31.4\\
    Ours  & \textbf{74.7} & \textbf{69.2} & \textbf{75.4} & \textbf{71.1} & \textbf{36.4} &\textbf{81.8} &\textbf{67.9} & \textbf{74.2}& \textbf{22.2}&\textbf{73.9}& \textbf{60.7}& \textbf{66.5}& \textbf{20.2}&\textbf{39.1}&\textbf{32.3}&\textbf{35.4}\\
\hline
    \end{tabular}
    \label{tab:main}
\end{table*}

\subsection{Comparison Methods}
To validate the model effectiveness, we compare with both existing classification-based  and generation-based methods. The first four are classification-based methods.

\begin{itemize}[leftmargin=*]
    \item \textbf{DARN}~\cite{huang2015cross}  is a Dual Attribute-aware Ranking Network originally used for retrieval feature learning. Same as in ~\cite{ma2019and}, we also only keep one stream for our task.
    \item \textbf{FashionNet}~\cite{liu2016deepfashion} is a pipeline-based model which simultaneously predicts landmarks and attributes. It consists of a global appearance branch, a local appearance branch and a pose branch.
    \item \textbf{HDF}~\cite{ma2019and}  extracts the fashion knowledge from social media posts. It unifies three tasks of occasion, person and clothing discovery from multiple modalities of images, texts and metadata.
    \item \textbf{ViLBERT}~\cite{lu2019vilbert} directly takes image text features as inputs and treats the task as a classification task. For occasion prediction, the input is the post image and text, and the output is the occasion category. While for fashion item information extraction, the image input is the fashion item box, the output is the type and appearance classes of the fashion item.
\end{itemize}
To further evaluate the effectiveness of our proposed captioning method, we also adopt the following generation-based baselines:
\begin{itemize}[leftmargin=*]
    \item \textbf{Oscar}~\cite{li2020oscar}. We utilize Oscar to generate the fashion knowledge tuples. Oscar is BERT-like and does not have an encoder-decoder structure. The model is pre-trained on several classification tasks and one generation task (COCOCaption). During fine-tuning, the words in the tuples serve as input and are masked randomly at the rate of 15\%. During inference, the generation process terminates when the model outputs the [STOP] token.
    \item \textbf{VL-Bart}~\cite{cho2021vlt5}. We also construct a baseline which uses the same pre-trained VL-Bart model as ours but without the proposed captioning method. Specifically, we directly employ the fashion knowledge tuples in the natural language form as the target sequence,  instead of transforming them into a natural language caption with the sentence transformation strategy.
\end{itemize}
\subsection{Experimental Setting}
In our experiment, we randomly split the dataset into the training, testing, and validation set with the percentage of 80\%, 10\%, 10\%. We conduct 5 runs for our experiment, each with a different random seed and report the average score. When comparing our method with existing models, since most of the existing  classification-based methods take the fashion item boxes as an input, we use an existing tool~\cite{zhang2016joint} to extract and predict all the person attributes in the post, following~\cite{ma2019and}. For each person box, we then use the same Faster-RCNN network mentioned in Section \ref{imageembedding} to extract all the fashion items. Our code is based on PyTorch and Huggingface Transformers ~\cite{wolf-etal-2020-transformers}. We use AdamW ~\cite{loshchilov2017decoupled}  with ($\beta_1$,$\beta_2$) = (0.9,0.999) and the  learning rate 1e-4 with 5\% linear warmup schedule. By default, each training process is run for 40 epochs.  We report the results from the top-20 fashion item boxes with the confidence score greater than 0.5 from the original extraction results.

We use several evaluation metrics in our experiment. At the fashion item level, we report the precision, recall, F1 rate of each fashion item tuple. A tuple prediction is counted as correct only when all the elements are the same as the ground-truth label. We also report the post-wise accuracy score, which is the probability of post  predictions our model got right. Except for that, to measure the semantic similarity between the generated caption and the transformed gold standard, we also employ some caption evaluation metrics including BLEU~\cite{papineni2002bleu} and METEOR~\cite{banerjee2005meteor}. 
\subsection{Main Experiment Results}
Table \ref{tab:main} shows the main experiment results of our model and the baseline models. Except for the overall fashion tuple prediction performance (``Overall''), we also report the performance of the occasion, fashion item category and appearance prediction for a more comprehensive comparison. We have the following observations:

First of all, error propagation is the main problem in the pipeline-based methods. The inaccurate prediction of the person boxes can lead to the fashion item information prediction errors, affecting both the category, appearance and overall performance. Secondly, for pipeline-based models, there remains a gap between the precision and recall rate in most tasks. For example, the  precision rate of HDF is 77.1 while the recall rate is 52.7 in the category prediction subtask. The gap mainly comes from the inaccurate fashion item box extraction, where many fashion items are not extracted or misextracted, thus leading to the low recall rate in the model result. 
In addition, some models take the dependencies between different fashion knowledge items into account (e.g. HDF), thus achieving better performance than those not (e.g. FashionNet). Moreover, it can also be observed that pre-trained models (e.g. ViLBERT) have a better performance than the non-pretrained models (e.g. DARN), showing the effectiveness of large pre-trained models in our task. 

Our model achieves the best overall performance among all the methods. Although Oscar outperforms our method in the occasion prediction subtask, where each post contains only one occasion label belonging to one of the six categories, which does not require the model to have a strong generation ability. Oscar has difficulty in generating the more complex fashion item information, getting only 5.5 of the overall accuracy score. In addition, our model gets further improved by transforming the original tuple-like fashion knowledge into natural language sentences. Compared with directly generating the fashion knowledge tuples (i.e. VL-Bart), the overall F1 score improves from 27.1 to 35.4. The result proves that generating the natural language caption helps the generation model capture the dependencies between different fashion knowledge aspects, thus resulting in a better prediction. 

To further study the role of different modalities in this task, we remove the post text and the image object tags respectively. 
Without post texts, the overall F1 score drops from 35.4 to 30.7. This result verifies that the post text contains rich fashion knowledge information with respect to where the post is located, who is in the post, and what the person is wearing. Besides, the image tags also play a vital role in our task, improving the overall performance from 31.4 to 35.4.  The reason is that some tags (e.g. woman, dress) can be aligned to the corresponding image regions and provide hints for the fashion knowledge such as the person gender and fashion item categories. The results verify that essential information is contained in different modalities of the post, which can be effectively captured by our model. 
\begin{table}[]
    \centering
    \caption{Performance comparison regarding different auxiliary tasks, where base denotes directly fine-tuning on our dataset without post-training. }
    \small
    \begin{tabular}{P{1.5cm}|P{1cm}|P{1cm}|P{1.2cm}|P{0.8cm}|P{0.8cm}}
    \hline
    Setting  &BLEU$_1$ &BLEU$_2$ & METEOR &Acc &F1 \\
    \hline
    Base & 69.77 & 64.51  &38.17 & 13.76 & 30.43  \\
    \hline
    +SRC   & 71.20 & 65.81 & 38.69& 15.08 & 31.79\\ 
    +ITM  & 71.90 & 66.05 &38.44 & 15.31 & 32.04\\
    +VQA   & 71.79 & 65.03  &38.81& 15.10 & 31.87 \\
    +ITM+VQA  & 72.49& 66.68  &38.90& 18.23 &  34.01\\
    +SRC+VQA  & 72.81 & 67.04 &38.79 & 17.97 &  32.56\\
    +SRC+ITM  & 73.46  & 67.61 &38.86& 18.54 & 33.87 \\
    \hline\hline
    Ours   & 75.40 & 69.29 & 39.80&20.22 & 35.43\\
    \hline
    \end{tabular}
    \label{tab:abl}
\end{table}
\subsection{Ablation Study}
To evaluate the effect of different auxiliary  tasks, we report the performance of our model with several variants. As shown in Table \ref{tab:abl}, we remove one or two auxiliary tasks at each time and report the corresponding accuracy and F1 scores. To further analyze the effects of those tasks on the generation results at the semantic level, BLEU and METEOR scores are also presented.

It can be noted that introducing auxiliary tasks improves the performance compared to directly fine-tuning the model on our dataset,  which enhances the model with fashion-related knowledge. Among the three tasks, ITM is the most beneficial to the performance improvement, which improves the BLEU$_1$ and BLEU$_2$ scores by 2.13 and 1.54 percent. The reason is that the captions of different images are constructed by the same transformation method and share similar structure, recognizing the right caption from the negative image-caption pairs helps the model understand the fashion knowledge elements better. Compared with other tasks, removing SRC (corresponds to ``+ITM+VQA'' in the table) has the least influence on the F1 score. The reason is that when masking the caption, some less important tokens which appear with high frequency are masked with an equal probability with tokens containing rich fashion knowledge. For example, in the caption sentence ``The woman wears a black upper'', token ``The'' has the same probability of being masked as the token ``upper''. 


\begin{figure}
    \centering
    \includegraphics[width=\linewidth]{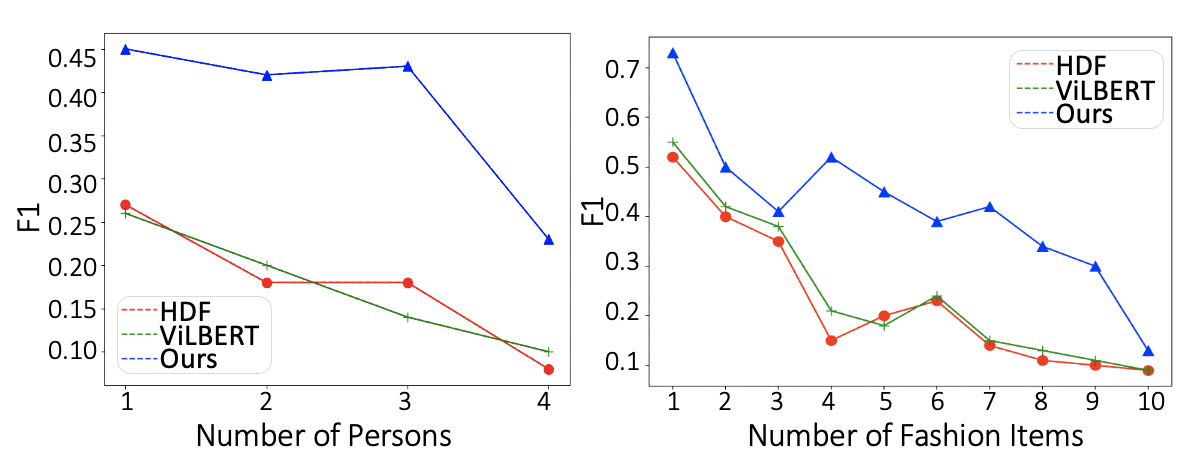}
    \caption{Performance comparison with respect to different person and fashion item numbers.}  
    \label{fig:an1}
    \vspace{-0.3cm}
\end{figure}

\begin{figure*}
    \centering
    \includegraphics[width=\linewidth]{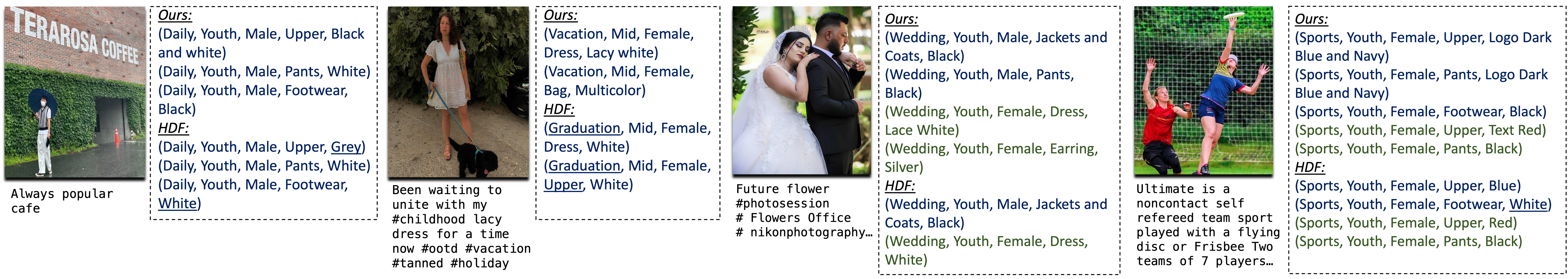}
    \caption{Real Case results of our model and HDF. Different color represents different person information. The  underline in the tuple denotes the errors in the prediction.}
    \label{fig:casestudy}
\end{figure*}

\subsection{Extensive Analysis}
\subsubsection{Performance under Different Person and Fashion Item Numbers} We analyze the performance of our model compared with the baseline models under different person and fashion item numbers, and plot the performance change in Figure \ref{fig:an1}. We can see that although the F1 score decreases for every method as the number of persons and fashion items grows, our model shows a greater advantage when it comes to the multi-person or multi-fashion-item setting. Specifically, when the number of persons and fashion items is small, both classification-based models and our model achieve a reasonable performance. However, as the case becomes more complicated, which means more persons are included in the image, traditional models often fail to extract all the fashion knowledge from the post. The performance gap between our method and a baseline HDF model reaches to the largest when there are 3 persons and 4 fashion items in the post image.
Such failure on the first place, may result from the error propagation for the pipeline-based method, which means the inaccurate extraction of person boxes may give rise to the wrong prediction of all the fashion items associated with that person. On the other place, compared with our model, most traditional models fail to capture the relationship between different fashion knowledge elements. For example, the occasion ``wedding'' can be related to a young woman wearing a lacy white dress and a young man wearing a black suit. Our model captures such correlation by generating a caption where the occasion and person attributes are generated first, which provides some prior hints for the upcoming fashion item knowledge generation.  

\subsubsection{Generative v.s. Discriminative Methods} 
As can be observed from Table~\ref{tab:main}, generative methods generally achieve better performance than the classification type methods. To further investigate such a phenomenon, we break down the testing dataset into three groups, namely common, rare, and unseen set. Specifically, we define the testing fashion tuples appearing more than 5 times in the training set as the common set, those contained in the training set but appear less than 5 times as the rare set. For fashion knowledge tuples that never appear in the training set, we denote them as the unseen set. Table \ref{tab:an2} shows the recall score of the three groups, which is the likelihood of the corresponding tuples being correctly predicted by the model. 

As shown in the table, our model improves upon the discriminative baselines across all the tuple categories. This improvement is more significant on the rare data, where the recall score improves by 15.01 percent compared with ViLBERT. The result demonstrates the effectiveness of generative models in the FKE  task, showing that when it comes to unfamiliar cases, generative models learn to describe the fashion items using the given knowledge compared to discriminative methods. What's more, by generating a natural language caption, the recall rate improves by 7.32 and 3.46 percent in rare and unseen cases, which proves that the interplay between image and text can be better captured compared with generating fashion knowledge tuples directly.

\begin{table}[]
    \centering
    \caption{Recall rate of generative and discriminative methods on different test categories.}
    \small

    \begin{tabular}{P{2cm}P{1.1cm}P{1.1cm}P{1.1cm}P{1.1cm}}
    \hline
    Method& Common &Rare & Unseen & Overall \\ \hline
      \textbf{Discriminative}   &  \\
       HDF  &  18.32 & 6.07 & 1.79 & 17.62\\
       ViLBERT  & 22.31 & 6.21 &1.97 &20.41 \\ \hline
      \textbf{Generative} & \\
        VL-Bart &26.21 &13.90 & 3.51& 21.85\\
        Ours & 34.63 & 21.22 & 6.97 & 32.28\\ \hline
    \end{tabular}
    \vspace{-0.3cm}
    \label{tab:an2}
\end{table}

\subsubsection{Caption Construction Analysis}
Our proposed sentence construction method transforms the original fashion knowledge tuples to a natural language caption for the sequence-to-sequence mapping. To verify the effectiveness of such  design, we also perform experiments based on different caption construction strategies and report the accuracy in Table \ref{rule}. We use three different caption construction rules in the experiment. Some rules are  designed to combine the fashion knowledge tuples in a less compact way (e.g. Rule 1 and 2). For example, we use one sentence to describe each fashion knowledge tuple respectively. We also design some rules where different fashion knowledge aspects are separated (e.g. Rule 3), where we follow the order of occasion first, person next, fashion item last when constructing the caption. 

To better demonstrate the algorithms of them, we use one example to illustrate. With the input of three fashion knowledge tuples \verb|(daily, P1, male, kid, upper, black), (daily, P1,| \verb|male, kid, pants, white), (daily, P2, female, old, dress,| \verb|blue)|, the outputs of them are as follows:
\begin{itemize}[leftmargin=*]
\item Rule 1 \emph{The first male kid wears a black upper in daily. The first male kid wears a white pants in daily. The second female old wears a blue dress in daily.}
\item Rule 2 \emph{The first male kid wears a black upper and a white pants in daily. The second female old wears a blue dress in daily.}
\item Rule 3 \emph{The occasion is daily. The person is a male kid and a female old. The first person wears a black upper and a white pants. The second person wears a blue dress.}
\item Ours \emph{The occasion is daily. The first male kid wears a black upper and a white pants. The second female old wears a blue dress.}
\end{itemize}
According to the results, our caption construction method achieves the best performance in all aspects. Rule 1 and 2 both put the occasion information at the end of each sentence. However, we find that it may pose a negative influence on the occasion prediction. Compared with Rule 3 where the person and fashion item information is separated, our method has a more compact form and helps to better capture the interplay between different aspects, improving the overall accuracy from 18.2 to 20.2.

\begin{table}[]
    \centering
    \caption{Analysis of different caption construction strategies. }
    \small
    \begin{tabular}{P{1.5cm}|P{1.2cm}|P{1.2cm}|P{1.2cm}|P{1.2cm}}\hline
        & Occ. & Cat. & App. & Overall\\ \hline
      \textbf{Rule 1}  & 72.5  & 36.0 & 21.8 & 18.0 \\
      \textbf{Rule 2}  & 73.6 &36.1 & 22.0& 19.1\\
      \textbf{Rule 3}  & 74.2 & 35.8& 22.1& 18.2\\ \hline
      \textbf{Ours} &  74.7 &36.4 & 22.2 &20.2\\ \hline
    \end{tabular}
    \label{rule}
\end{table}

\subsection{Case Study}
We use some real cases to compare the performance of our model with the HDF model in a more vivid way. As shown in Figure  \ref{fig:casestudy}, there are more errors in the HDF extraction results compared with our model. For example, the appearance of the upper in the first case is misclassified as grey. In addition, our model captures the interplay between the image and text information better. For example, in the second case, the post text ``\textit{lacy dress}'' corresponds to the dress in the image and the hashtag indicates that the occasion should be vacation. 
What's more, our model provides more comprehensive results. For example, in the third case, the HDF model fails to extract the less obvious earring information in the image and also ignores the pants the man wears. 
Also concerning the appearance of the fashion items, our model outputs better description against the HDF model. As shown in the third case, our model describes the dress of the woman as ``\textit{lacy white}'', while the HDF model only classifies the dress as ``\textit{white}''. 

\section{Conclusion}
We investigate social media based FKE task. For a social media post consisting of an image and text, we aim to elicit the occasion, person attributes, and fashion item information from the post. Specifically, we formulate this task as a captioning problem and transform the fashion knowledge tuples into a natural language caption. We also design several auxiliary tasks before captioning to warm-up the model with task-specific knowledge. Since no existing dataset can be directly adapted to our task, we contribute a large-scale dataset with manual annotation. Extensive experiments are conducted to demonstrate the effectiveness of our model.

\appendix




\bibliographystyle{ACM-Reference-Format}
\bibliography{fke}


\begin{thebibliography}{50}


\ifx \showCODEN    \undefined \def \showCODEN     #1{\unskip}     \fi
\ifx \showDOI      \undefined \def \showDOI       #1{#1}\fi
\ifx \showISBNx    \undefined \def \showISBNx     #1{\unskip}     \fi
\ifx \showISBNxiii \undefined \def \showISBNxiii  #1{\unskip}     \fi
\ifx \showISSN     \undefined \def \showISSN      #1{\unskip}     \fi
\ifx \showLCCN     \undefined \def \showLCCN      #1{\unskip}     \fi
\ifx \shownote     \undefined \def \shownote      #1{#1}          \fi
\ifx \showarticletitle \undefined \def \showarticletitle #1{#1}   \fi
\ifx \showURL      \undefined \def \showURL       {\relax}        \fi
\providecommand\bibfield[2]{#2}
\providecommand\bibinfo[2]{#2}
\providecommand\natexlab[1]{#1}
\providecommand\showeprint[2][]{arXiv:#2}

\bibitem[\protect\citeauthoryear{Antol, Agrawal, Lu, Mitchell, Batra, Zitnick, and Parikh}{Antol et~al\mbox{.}}{2015}]%
        {antol2015vqa}
\bibfield{author}{\bibinfo{person}{Stanislaw Antol}, \bibinfo{person}{Aishwarya Agrawal}, \bibinfo{person}{Jiasen Lu}, \bibinfo{person}{Margaret Mitchell}, \bibinfo{person}{Dhruv Batra}, \bibinfo{person}{C~Lawrence Zitnick}, {and} \bibinfo{person}{Devi Parikh}.} \bibinfo{year}{2015}\natexlab{}.
\newblock \showarticletitle{Vqa: Visual question answering}. In \bibinfo{booktitle}{\emph{Proceedings of the IEEE international conference on computer vision}}. \bibinfo{pages}{2425--2433}.
\newblock


\bibitem[\protect\citeauthoryear{Banerjee and Lavie}{Banerjee and Lavie}{2005}]%
        {banerjee2005meteor}
\bibfield{author}{\bibinfo{person}{Satanjeev Banerjee} {and} \bibinfo{person}{Alon Lavie}.} \bibinfo{year}{2005}\natexlab{}.
\newblock \showarticletitle{METEOR: An automatic metric for MT evaluation with improved correlation with human judgments}. In \bibinfo{booktitle}{\emph{Proceedings of the acl workshop on intrinsic and extrinsic evaluation measures for machine translation and/or summarization}}. \bibinfo{pages}{65--72}.
\newblock


\bibitem[\protect\citeauthoryear{Brown, Mann, Ryder, Subbiah, Kaplan, Dhariwal, Neelakantan, Shyam, Sastry, Askell, et~al\mbox{.}}{Brown et~al\mbox{.}}{2020}]%
        {brown2020language}
\bibfield{author}{\bibinfo{person}{Tom~B Brown}, \bibinfo{person}{Benjamin Mann}, \bibinfo{person}{Nick Ryder}, \bibinfo{person}{Melanie Subbiah}, \bibinfo{person}{Jared Kaplan}, \bibinfo{person}{Prafulla Dhariwal}, \bibinfo{person}{Arvind Neelakantan}, \bibinfo{person}{Pranav Shyam}, \bibinfo{person}{Girish Sastry}, \bibinfo{person}{Amanda Askell}, {et~al\mbox{.}}} \bibinfo{year}{2020}\natexlab{}.
\newblock \showarticletitle{Language models are few-shot learners}.
\newblock \bibinfo{journal}{\emph{arXiv preprint arXiv:2005.14165}} (\bibinfo{year}{2020}).
\newblock


\bibitem[\protect\citeauthoryear{Chen, Gallagher, and Girod}{Chen et~al\mbox{.}}{2012}]%
        {chen2012describing}
\bibfield{author}{\bibinfo{person}{Huizhong Chen}, \bibinfo{person}{Andrew Gallagher}, {and} \bibinfo{person}{Bernd Girod}.} \bibinfo{year}{2012}\natexlab{}.
\newblock \showarticletitle{Describing clothing by semantic attributes}. In \bibinfo{booktitle}{\emph{European conference on computer vision}}. Springer, \bibinfo{pages}{609--623}.
\newblock


\bibitem[\protect\citeauthoryear{Cho, Lei, Tan, and Bansal}{Cho et~al\mbox{.}}{2021}]%
        {cho2021vlt5}
\bibfield{author}{\bibinfo{person}{Jaemin Cho}, \bibinfo{person}{Jie Lei}, \bibinfo{person}{Hao Tan}, {and} \bibinfo{person}{Mohit Bansal}.} \bibinfo{year}{2021}\natexlab{}.
\newblock \showarticletitle{Unifying Vision-and-Language Tasks via Text Generation}. In \bibinfo{booktitle}{\emph{ICML}}.
\newblock


\bibitem[\protect\citeauthoryear{Clark, Luong, Le, and Manning}{Clark et~al\mbox{.}}{2020}]%
        {clark2020electra}
\bibfield{author}{\bibinfo{person}{Kevin Clark}, \bibinfo{person}{Minh-Thang Luong}, \bibinfo{person}{Quoc~V. Le}, {and} \bibinfo{person}{Christopher~D. Manning}.} \bibinfo{year}{2020}\natexlab{}.
\newblock \showarticletitle{{ELECTRA}: Pre-training Text Encoders as Discriminators Rather Than Generators}. In \bibinfo{booktitle}{\emph{ICLR}}.
\newblock


\bibitem[\protect\citeauthoryear{Deng, Li, Ding, and Lam}{Deng et~al\mbox{.}}{2022}]%
        {tkde-crs}
\bibfield{author}{\bibinfo{person}{Yang Deng}, \bibinfo{person}{Yaliang Li}, \bibinfo{person}{Bolin Ding}, {and} \bibinfo{person}{Wai Lam}.} \bibinfo{year}{2022}\natexlab{}.
\newblock \showarticletitle{Leveraging Long Short-Term User Preference in Conversational Recommendation Via Multi-Agent Reinforcement Learning}.
\newblock \bibinfo{journal}{\emph{IEEE Transactions on Knowledge and Data Engineering}} (\bibinfo{year}{2022}).
\newblock


\bibitem[\protect\citeauthoryear{Deng, Li, Sun, Ding, and Lam}{Deng et~al\mbox{.}}{2021}]%
        {sigir21-crs}
\bibfield{author}{\bibinfo{person}{Yang Deng}, \bibinfo{person}{Yaliang Li}, \bibinfo{person}{Fei Sun}, \bibinfo{person}{Bolin Ding}, {and} \bibinfo{person}{Wai Lam}.} \bibinfo{year}{2021}\natexlab{}.
\newblock \showarticletitle{Unified Conversational Recommendation Policy Learning via Graph-based Reinforcement Learning}. In \bibinfo{booktitle}{\emph{{SIGIR} 2021}}. \bibinfo{pages}{1431--1441}.
\newblock


\bibitem[\protect\citeauthoryear{Devlin, Chang, Lee, and Toutanova}{Devlin et~al\mbox{.}}{2018}]%
        {devlin2018bert}
\bibfield{author}{\bibinfo{person}{Jacob Devlin}, \bibinfo{person}{Ming-Wei Chang}, \bibinfo{person}{Kenton Lee}, {and} \bibinfo{person}{Kristina Toutanova}.} \bibinfo{year}{2018}\natexlab{}.
\newblock \showarticletitle{Bert: Pre-training of deep bidirectional transformers for language understanding}.
\newblock \bibinfo{journal}{\emph{arXiv preprint arXiv:1810.04805}} (\bibinfo{year}{2018}).
\newblock


\bibitem[\protect\citeauthoryear{Ding, Yang, Hong, Zheng, Zhou, Yin, Lin, Zou, Shao, Yang, and Tang}{Ding et~al\mbox{.}}{2021}]%
        {ding2021cogview}
\bibfield{author}{\bibinfo{person}{Ming Ding}, \bibinfo{person}{Zhuoyi Yang}, \bibinfo{person}{Wenyi Hong}, \bibinfo{person}{Wendi Zheng}, \bibinfo{person}{Chang Zhou}, \bibinfo{person}{Da Yin}, \bibinfo{person}{Junyang Lin}, \bibinfo{person}{Xu Zou}, \bibinfo{person}{Zhou Shao}, \bibinfo{person}{Hongxia Yang}, {and} \bibinfo{person}{Jie Tang}.} \bibinfo{year}{2021}\natexlab{}.
\newblock \showarticletitle{CogView: Mastering Text-to-Image Generation via Transformers}.
\newblock \bibinfo{journal}{\emph{arXiv preprint arXiv:2105.13290}} (\bibinfo{year}{2021}).
\newblock


\bibitem[\protect\citeauthoryear{Hu, Yi, and Davis}{Hu et~al\mbox{.}}{2015}]%
        {hu2015collaborative}
\bibfield{author}{\bibinfo{person}{Yang Hu}, \bibinfo{person}{Xi Yi}, {and} \bibinfo{person}{Larry~S Davis}.} \bibinfo{year}{2015}\natexlab{}.
\newblock \showarticletitle{Collaborative fashion recommendation: A functional tensor factorization approach}. In \bibinfo{booktitle}{\emph{Proceedings of the 23rd ACM international conference on Multimedia}}. \bibinfo{pages}{129--138}.
\newblock


\bibitem[\protect\citeauthoryear{Huang, Feris, Chen, and Yan}{Huang et~al\mbox{.}}{2015}]%
        {huang2015cross}
\bibfield{author}{\bibinfo{person}{Junshi Huang}, \bibinfo{person}{Rogerio~S Feris}, \bibinfo{person}{Qiang Chen}, {and} \bibinfo{person}{Shuicheng Yan}.} \bibinfo{year}{2015}\natexlab{}.
\newblock \showarticletitle{Cross-domain image retrieval with a dual attribute-aware ranking network}. In \bibinfo{booktitle}{\emph{Proceedings of the IEEE international conference on computer vision}}. \bibinfo{pages}{1062--1070}.
\newblock


\bibitem[\protect\citeauthoryear{Huo, Zhang, Liu, Lu, Gao, Yang, Wen, Zhang, Xu, Zheng, et~al\mbox{.}}{Huo et~al\mbox{.}}{2021}]%
        {huo2021wenlan}
\bibfield{author}{\bibinfo{person}{Yuqi Huo}, \bibinfo{person}{Manli Zhang}, \bibinfo{person}{Guangzhen Liu}, \bibinfo{person}{Haoyu Lu}, \bibinfo{person}{Yizhao Gao}, \bibinfo{person}{Guoxing Yang}, \bibinfo{person}{Jingyuan Wen}, \bibinfo{person}{Heng Zhang}, \bibinfo{person}{Baogui Xu}, \bibinfo{person}{Weihao Zheng}, {et~al\mbox{.}}} \bibinfo{year}{2021}\natexlab{}.
\newblock \showarticletitle{WenLan: Bridging vision and language by large-scale multi-modal pre-training}.
\newblock \bibinfo{journal}{\emph{arXiv preprint arXiv:2103.06561}} (\bibinfo{year}{2021}).
\newblock


\bibitem[\protect\citeauthoryear{Jia, Zhou, Shi, and Hariharan}{Jia et~al\mbox{.}}{2018}]%
        {jia2018deep}
\bibfield{author}{\bibinfo{person}{Menglin Jia}, \bibinfo{person}{Yichen Zhou}, \bibinfo{person}{Mengyun Shi}, {and} \bibinfo{person}{Bharath Hariharan}.} \bibinfo{year}{2018}\natexlab{}.
\newblock \showarticletitle{A deep-learning-based fashion attributes detection model}.
\newblock \bibinfo{journal}{\emph{arXiv preprint arXiv:1810.10148}} (\bibinfo{year}{2018}).
\newblock


\bibitem[\protect\citeauthoryear{Kalantidis, Kennedy, and Li}{Kalantidis et~al\mbox{.}}{2013}]%
        {kalantidis2013getting}
\bibfield{author}{\bibinfo{person}{Yannis Kalantidis}, \bibinfo{person}{Lyndon Kennedy}, {and} \bibinfo{person}{Li-Jia Li}.} \bibinfo{year}{2013}\natexlab{}.
\newblock \showarticletitle{Getting the look: clothing recognition and segmentation for automatic product suggestions in everyday photos}. In \bibinfo{booktitle}{\emph{Proceedings of the 3rd ACM conference on International conference on multimedia retrieval}}. \bibinfo{pages}{105--112}.
\newblock


\bibitem[\protect\citeauthoryear{Kang, Fang, Wang, and McAuley}{Kang et~al\mbox{.}}{2017}]%
        {kang2017visually}
\bibfield{author}{\bibinfo{person}{Wang-Cheng Kang}, \bibinfo{person}{Chen Fang}, \bibinfo{person}{Zhaowen Wang}, {and} \bibinfo{person}{Julian McAuley}.} \bibinfo{year}{2017}\natexlab{}.
\newblock \showarticletitle{Visually-aware fashion recommendation and design with generative image models}. In \bibinfo{booktitle}{\emph{2017 IEEE International Conference on Data Mining (ICDM)}}. IEEE, \bibinfo{pages}{207--216}.
\newblock


\bibitem[\protect\citeauthoryear{Lewis, Liu, Goyal, Ghazvininejad, Mohamed, Levy, Stoyanov, and Zettlemoyer}{Lewis et~al\mbox{.}}{2019}]%
        {lewis2019bart}
\bibfield{author}{\bibinfo{person}{Mike Lewis}, \bibinfo{person}{Yinhan Liu}, \bibinfo{person}{Naman Goyal}, \bibinfo{person}{Marjan Ghazvininejad}, \bibinfo{person}{Abdelrahman Mohamed}, \bibinfo{person}{Omer Levy}, \bibinfo{person}{Ves Stoyanov}, {and} \bibinfo{person}{Luke Zettlemoyer}.} \bibinfo{year}{2019}\natexlab{}.
\newblock \showarticletitle{Bart: Denoising sequence-to-sequence pre-training for natural language generation, translation, and comprehension}.
\newblock \bibinfo{journal}{\emph{arXiv preprint arXiv:1910.13461}} (\bibinfo{year}{2019}).
\newblock


\bibitem[\protect\citeauthoryear{Li, Chen, Cheng, Gan, Yu, and Liu}{Li et~al\mbox{.}}{2020a}]%
        {li2020hero}
\bibfield{author}{\bibinfo{person}{Linjie Li}, \bibinfo{person}{Yen-Chun Chen}, \bibinfo{person}{Yu Cheng}, \bibinfo{person}{Zhe Gan}, \bibinfo{person}{Licheng Yu}, {and} \bibinfo{person}{Jingjing Liu}.} \bibinfo{year}{2020}\natexlab{a}.
\newblock \showarticletitle{HERO: Hierarchical Encoder for Video+ Language Omni-representation Pre-training}. In \bibinfo{booktitle}{\emph{EMNLP}}.
\newblock


\bibitem[\protect\citeauthoryear{Li, Yin, Li, Hu, Zhang, Zhang, Wang, Hu, Dong, Wei, Choi, and Gao}{Li et~al\mbox{.}}{2020b}]%
        {li2020oscar}
\bibfield{author}{\bibinfo{person}{Xiujun Li}, \bibinfo{person}{Xi Yin}, \bibinfo{person}{Chunyuan Li}, \bibinfo{person}{Xiaowei Hu}, \bibinfo{person}{Pengchuan Zhang}, \bibinfo{person}{Lei Zhang}, \bibinfo{person}{Lijuan Wang}, \bibinfo{person}{Houdong Hu}, \bibinfo{person}{Li Dong}, \bibinfo{person}{Furu Wei}, \bibinfo{person}{Yejin Choi}, {and} \bibinfo{person}{Jianfeng Gao}.} \bibinfo{year}{2020}\natexlab{b}.
\newblock \showarticletitle{Oscar: Object-Semantics Aligned Pre-training for Vision-Language Tasks}.
\newblock \bibinfo{journal}{\emph{ECCV 2020}} (\bibinfo{year}{2020}).
\newblock


\bibitem[\protect\citeauthoryear{Liu and Lu}{Liu and Lu}{2018}]%
        {liu2018deep}
\bibfield{author}{\bibinfo{person}{Jingyuan Liu} {and} \bibinfo{person}{Hong Lu}.} \bibinfo{year}{2018}\natexlab{}.
\newblock \showarticletitle{Deep Fashion Analysis with Feature Map Upsampling and Landmark-Driven Attention}. In \bibinfo{booktitle}{\emph{European Conference on Computer Vision}}. Springer, \bibinfo{pages}{30--36}.
\newblock


\bibitem[\protect\citeauthoryear{Liu, Luo, Qiu, Wang, and Tang}{Liu et~al\mbox{.}}{2016}]%
        {liu2016deepfashion}
\bibfield{author}{\bibinfo{person}{Ziwei Liu}, \bibinfo{person}{Ping Luo}, \bibinfo{person}{Shi Qiu}, \bibinfo{person}{Xiaogang Wang}, {and} \bibinfo{person}{Xiaoou Tang}.} \bibinfo{year}{2016}\natexlab{}.
\newblock \showarticletitle{DeepFashion: Powering Robust Clothes Recognition and Retrieval with Rich Annotations}. In \bibinfo{booktitle}{\emph{Proceedings of IEEE Conference on Computer Vision and Pattern Recognition (CVPR)}}.
\newblock


\bibitem[\protect\citeauthoryear{Loshchilov and Hutter}{Loshchilov and Hutter}{2017}]%
        {loshchilov2017decoupled}
\bibfield{author}{\bibinfo{person}{Ilya Loshchilov} {and} \bibinfo{person}{Frank Hutter}.} \bibinfo{year}{2017}\natexlab{}.
\newblock \showarticletitle{Decoupled weight decay regularization}.
\newblock \bibinfo{journal}{\emph{arXiv preprint arXiv:1711.05101}} (\bibinfo{year}{2017}).
\newblock


\bibitem[\protect\citeauthoryear{Lu, Batra, Parikh, and Lee}{Lu et~al\mbox{.}}{2019}]%
        {lu2019vilbert}
\bibfield{author}{\bibinfo{person}{Jiasen Lu}, \bibinfo{person}{Dhruv Batra}, \bibinfo{person}{Devi Parikh}, {and} \bibinfo{person}{Stefan Lee}.} \bibinfo{year}{2019}\natexlab{}.
\newblock \showarticletitle{Vilbert: Pretraining task-agnostic visiolinguistic representations for vision-and-language tasks}. In \bibinfo{booktitle}{\emph{Advances in Neural Information Processing Systems}}. \bibinfo{pages}{13--23}.
\newblock


\bibitem[\protect\citeauthoryear{Ma, Ding, Yang, Liao, Wong, and Chua}{Ma et~al\mbox{.}}{2020}]%
        {ma2020knowledge}
\bibfield{author}{\bibinfo{person}{Yunshan Ma}, \bibinfo{person}{Yujuan Ding}, \bibinfo{person}{Xun Yang}, \bibinfo{person}{Lizi Liao}, \bibinfo{person}{Wai~Keung Wong}, {and} \bibinfo{person}{Tat-Seng Chua}.} \bibinfo{year}{2020}\natexlab{}.
\newblock \showarticletitle{Knowledge enhanced neural fashion trend forecasting}. In \bibinfo{booktitle}{\emph{Proceedings of the 2020 International Conference on Multimedia Retrieval}}. \bibinfo{pages}{82--90}.
\newblock


\bibitem[\protect\citeauthoryear{Ma, Yang, Liao, Cao, and Chua}{Ma et~al\mbox{.}}{2019}]%
        {ma2019and}
\bibfield{author}{\bibinfo{person}{Yunshan Ma}, \bibinfo{person}{Xun Yang}, \bibinfo{person}{Lizi Liao}, \bibinfo{person}{Yixin Cao}, {and} \bibinfo{person}{Tat-Seng Chua}.} \bibinfo{year}{2019}\natexlab{}.
\newblock \showarticletitle{Who, where, and what to wear? Extracting fashion knowledge from social media}. In \bibinfo{booktitle}{\emph{Proceedings of the 27th ACM International Conference on Multimedia}}. \bibinfo{pages}{257--265}.
\newblock


\bibitem[\protect\citeauthoryear{Mall, Matzen, Hariharan, Snavely, and Bala}{Mall et~al\mbox{.}}{2019}]%
        {mall2019geostyle}
\bibfield{author}{\bibinfo{person}{Utkarsh Mall}, \bibinfo{person}{Kevin Matzen}, \bibinfo{person}{Bharath Hariharan}, \bibinfo{person}{Noah Snavely}, {and} \bibinfo{person}{Kavita Bala}.} \bibinfo{year}{2019}\natexlab{}.
\newblock \showarticletitle{Geostyle: Discovering fashion trends and events}. In \bibinfo{booktitle}{\emph{Proceedings of the IEEE/CVF International Conference on Computer Vision}}. \bibinfo{pages}{411--420}.
\newblock


\bibitem[\protect\citeauthoryear{Matzen, Bala, and Snavely}{Matzen et~al\mbox{.}}{2017}]%
        {matzen2017streetstyle}
\bibfield{author}{\bibinfo{person}{Kevin Matzen}, \bibinfo{person}{Kavita Bala}, {and} \bibinfo{person}{Noah Snavely}.} \bibinfo{year}{2017}\natexlab{}.
\newblock \showarticletitle{Streetstyle: Exploring world-wide clothing styles from millions of photos}.
\newblock \bibinfo{journal}{\emph{arXiv preprint arXiv:1706.01869}} (\bibinfo{year}{2017}).
\newblock


\bibitem[\protect\citeauthoryear{Miech, Alayrac, Smaira, Laptev, Sivic, and Zisserman}{Miech et~al\mbox{.}}{2020}]%
        {miech19endtoend}
\bibfield{author}{\bibinfo{person}{Antoine Miech}, \bibinfo{person}{Jean-Baptiste Alayrac}, \bibinfo{person}{Lucas Smaira}, \bibinfo{person}{Ivan Laptev}, \bibinfo{person}{Josef Sivic}, {and} \bibinfo{person}{Andrew Zisserman}.} \bibinfo{year}{2020}\natexlab{}.
\newblock \showarticletitle{{E}nd-to-{E}nd {L}earning of {V}isual {R}epresentations from {U}ncurated {I}nstructional {V}ideos}. In \bibinfo{booktitle}{\emph{CVPR}}.
\newblock


\bibitem[\protect\citeauthoryear{Miech, Zhukov, Alayrac, Tapaswi, Laptev, and Sivic}{Miech et~al\mbox{.}}{2019}]%
        {miech19howto100m}
\bibfield{author}{\bibinfo{person}{Antoine Miech}, \bibinfo{person}{Dimitri Zhukov}, \bibinfo{person}{Jean-Baptiste Alayrac}, \bibinfo{person}{Makarand Tapaswi}, \bibinfo{person}{Ivan Laptev}, {and} \bibinfo{person}{Josef Sivic}.} \bibinfo{year}{2019}\natexlab{}.
\newblock \showarticletitle{How{T}o100{M}: {L}earning a {T}ext-{V}ideo {E}mbedding by {W}atching {H}undred {M}illion {N}arrated {V}ideo {C}lips}. In \bibinfo{booktitle}{\emph{ICCV}}.
\newblock


\bibitem[\protect\citeauthoryear{Papineni, Roukos, Ward, and Zhu}{Papineni et~al\mbox{.}}{2002}]%
        {papineni2002bleu}
\bibfield{author}{\bibinfo{person}{Kishore Papineni}, \bibinfo{person}{Salim Roukos}, \bibinfo{person}{Todd Ward}, {and} \bibinfo{person}{Wei-Jing Zhu}.} \bibinfo{year}{2002}\natexlab{}.
\newblock \showarticletitle{Bleu: a method for automatic evaluation of machine translation}. In \bibinfo{booktitle}{\emph{Proceedings of the 40th annual meeting of the Association for Computational Linguistics}}. \bibinfo{pages}{311--318}.
\newblock


\bibitem[\protect\citeauthoryear{Radford, Sutskever, Jong Wook~Kim, and Agarwal}{Radford et~al\mbox{.}}{2021}]%
        {unpublished2021clip}
\bibfield{author}{\bibinfo{person}{Alec Radford}, \bibinfo{person}{Ilya Sutskever}, \bibinfo{person}{Gretchen~Krueger Jong Wook~Kim}, {and} \bibinfo{person}{Sandhini Agarwal}.} \bibinfo{year}{2021}\natexlab{}.
\newblock \bibinfo{title}{CLIP: Connecting Text and Images}.
\newblock
\newblock


\bibitem[\protect\citeauthoryear{Raffel, Shazeer, Roberts, Lee, Narang, Matena, Zhou, Li, and Liu}{Raffel et~al\mbox{.}}{2019}]%
        {raffel2019exploring}
\bibfield{author}{\bibinfo{person}{Colin Raffel}, \bibinfo{person}{Noam Shazeer}, \bibinfo{person}{Adam Roberts}, \bibinfo{person}{Katherine Lee}, \bibinfo{person}{Sharan Narang}, \bibinfo{person}{Michael Matena}, \bibinfo{person}{Yanqi Zhou}, \bibinfo{person}{Wei Li}, {and} \bibinfo{person}{Peter~J Liu}.} \bibinfo{year}{2019}\natexlab{}.
\newblock \showarticletitle{Exploring the limits of transfer learning with a unified text-to-text transformer}.
\newblock \bibinfo{journal}{\emph{arXiv preprint arXiv:1910.10683}} (\bibinfo{year}{2019}).
\newblock


\bibitem[\protect\citeauthoryear{Ramesh, Pavlov, Goh, Gray, Voss, Radford, Chen, and Sutskever}{Ramesh et~al\mbox{.}}{2021}]%
        {ramesh2021zeroshot}
\bibfield{author}{\bibinfo{person}{Aditya Ramesh}, \bibinfo{person}{Mikhail Pavlov}, \bibinfo{person}{Gabriel Goh}, \bibinfo{person}{Scott Gray}, \bibinfo{person}{Chelsea Voss}, \bibinfo{person}{Alec Radford}, \bibinfo{person}{Mark Chen}, {and} \bibinfo{person}{Ilya Sutskever}.} \bibinfo{year}{2021}\natexlab{}.
\newblock \bibinfo{title}{Zero-Shot Text-to-Image Generation}.
\newblock
\newblock
\showeprint[arxiv]{2102.12092}~[cs.CV]


\bibitem[\protect\citeauthoryear{Ren, He, Girshick, and Sun}{Ren et~al\mbox{.}}{2015}]%
        {renNIPS15fasterrcnn}
\bibfield{author}{\bibinfo{person}{Shaoqing Ren}, \bibinfo{person}{Kaiming He}, \bibinfo{person}{Ross Girshick}, {and} \bibinfo{person}{Jian Sun}.} \bibinfo{year}{2015}\natexlab{}.
\newblock \showarticletitle{Faster {R-CNN}: Towards Real-Time Object Detection with Region Proposal Networks}. In \bibinfo{booktitle}{\emph{Advances in Neural Information Processing Systems ({NIPS})}}.
\newblock


\bibitem[\protect\citeauthoryear{Su, Zhu, Cao, Li, Lu, Wei, and Dai}{Su et~al\mbox{.}}{2020}]%
        {Su2020VL-BERT:}
\bibfield{author}{\bibinfo{person}{Weijie Su}, \bibinfo{person}{Xizhou Zhu}, \bibinfo{person}{Yue Cao}, \bibinfo{person}{Bin Li}, \bibinfo{person}{Lewei Lu}, \bibinfo{person}{Furu Wei}, {and} \bibinfo{person}{Jifeng Dai}.} \bibinfo{year}{2020}\natexlab{}.
\newblock \showarticletitle{VL-BERT: Pre-training of Generic Visual-Linguistic Representations}. In \bibinfo{booktitle}{\emph{International Conference on Learning Representations}}.
\newblock


\bibitem[\protect\citeauthoryear{Sun, Myers, Vondrick, Murphy, and Schmid}{Sun et~al\mbox{.}}{2019}]%
        {DBLP:conf/iccv/SunMV0S19}
\bibfield{author}{\bibinfo{person}{Chen Sun}, \bibinfo{person}{Austin Myers}, \bibinfo{person}{Carl Vondrick}, \bibinfo{person}{Kevin Murphy}, {and} \bibinfo{person}{Cordelia Schmid}.} \bibinfo{year}{2019}\natexlab{}.
\newblock \showarticletitle{VideoBERT: {A} Joint Model for Video and Language Representation Learning}. In \bibinfo{booktitle}{\emph{2019 {IEEE/CVF} International Conference on Computer Vision, {ICCV} 2019, Seoul, Korea (South), October 27 - November 2, 2019}}. \bibinfo{publisher}{{IEEE}}, \bibinfo{pages}{7463--7472}.
\newblock


\bibitem[\protect\citeauthoryear{Vaswani, Shazeer, Parmar, Uszkoreit, Jones, Gomez, Kaiser, and Polosukhin}{Vaswani et~al\mbox{.}}{2017}]%
        {vaswani2017attention}
\bibfield{author}{\bibinfo{person}{Ashish Vaswani}, \bibinfo{person}{Noam Shazeer}, \bibinfo{person}{Niki Parmar}, \bibinfo{person}{Jakob Uszkoreit}, \bibinfo{person}{Llion Jones}, \bibinfo{person}{Aidan~N Gomez}, \bibinfo{person}{{\L}ukasz Kaiser}, {and} \bibinfo{person}{Illia Polosukhin}.} \bibinfo{year}{2017}\natexlab{}.
\newblock \showarticletitle{Attention is all you need}. In \bibinfo{booktitle}{\emph{Advances in neural information processing systems}}. \bibinfo{pages}{5998--6008}.
\newblock


\bibitem[\protect\citeauthoryear{Vinyals, Toshev, Bengio, and Erhan}{Vinyals et~al\mbox{.}}{2015}]%
        {vinyals2015show}
\bibfield{author}{\bibinfo{person}{Oriol Vinyals}, \bibinfo{person}{Alexander Toshev}, \bibinfo{person}{Samy Bengio}, {and} \bibinfo{person}{Dumitru Erhan}.} \bibinfo{year}{2015}\natexlab{}.
\newblock \showarticletitle{Show and tell: A neural image caption generator}. In \bibinfo{booktitle}{\emph{Proceedings of the IEEE conference on computer vision and pattern recognition}}. \bibinfo{pages}{3156--3164}.
\newblock


\bibitem[\protect\citeauthoryear{Wang, Xu, Shen, and Zhu}{Wang et~al\mbox{.}}{2018}]%
        {wang2018attentive}
\bibfield{author}{\bibinfo{person}{Wenguan Wang}, \bibinfo{person}{Yuanlu Xu}, \bibinfo{person}{Jianbing Shen}, {and} \bibinfo{person}{Song-Chun Zhu}.} \bibinfo{year}{2018}\natexlab{}.
\newblock \showarticletitle{Attentive fashion grammar network for fashion landmark detection and clothing category classification}. In \bibinfo{booktitle}{\emph{Proceedings of the IEEE Conference on Computer Vision and Pattern Recognition}}. \bibinfo{pages}{4271--4280}.
\newblock


\bibitem[\protect\citeauthoryear{Wolf, Debut, Sanh, Chaumond, Delangue, Moi, Cistac, Rault, Louf, Funtowicz, Davison, Shleifer, von Platen, Ma, Jernite, Plu, Xu, Scao, Gugger, Drame, Lhoest, and Rush}{Wolf et~al\mbox{.}}{2020}]%
        {wolf-etal-2020-transformers}
\bibfield{author}{\bibinfo{person}{Thomas Wolf}, \bibinfo{person}{Lysandre Debut}, \bibinfo{person}{Victor Sanh}, \bibinfo{person}{Julien Chaumond}, \bibinfo{person}{Clement Delangue}, \bibinfo{person}{Anthony Moi}, \bibinfo{person}{Pierric Cistac}, \bibinfo{person}{Tim Rault}, \bibinfo{person}{Rémi Louf}, \bibinfo{person}{Morgan Funtowicz}, \bibinfo{person}{Joe Davison}, \bibinfo{person}{Sam Shleifer}, \bibinfo{person}{Patrick von Platen}, \bibinfo{person}{Clara Ma}, \bibinfo{person}{Yacine Jernite}, \bibinfo{person}{Julien Plu}, \bibinfo{person}{Canwen Xu}, \bibinfo{person}{Teven~Le Scao}, \bibinfo{person}{Sylvain Gugger}, \bibinfo{person}{Mariama Drame}, \bibinfo{person}{Quentin Lhoest}, {and} \bibinfo{person}{Alexander~M. Rush}.} \bibinfo{year}{2020}\natexlab{}.
\newblock \showarticletitle{Transformers: State-of-the-Art Natural Language Processing}. In \bibinfo{booktitle}{\emph{Proceedings of the 2020 Conference on Empirical Methods in Natural Language Processing: System Demonstrations}}. \bibinfo{publisher}{Association for Computational Linguistics}, \bibinfo{address}{Online}, \bibinfo{pages}{38--45}.
\newblock


\bibitem[\protect\citeauthoryear{Wu, Gao, Guo, Al-Halah, Rennie, Grauman, and Feris}{Wu et~al\mbox{.}}{2021}]%
        {wu2021fashion}
\bibfield{author}{\bibinfo{person}{Hui Wu}, \bibinfo{person}{Yupeng Gao}, \bibinfo{person}{Xiaoxiao Guo}, \bibinfo{person}{Ziad Al-Halah}, \bibinfo{person}{Steven Rennie}, \bibinfo{person}{Kristen Grauman}, {and} \bibinfo{person}{Rogerio Feris}.} \bibinfo{year}{2021}\natexlab{}.
\newblock \showarticletitle{Fashion iq: A new dataset towards retrieving images by natural language feedback}. In \bibinfo{booktitle}{\emph{Proceedings of the IEEE/CVF Conference on Computer Vision and Pattern Recognition}}. \bibinfo{pages}{11307--11317}.
\newblock


\bibitem[\protect\citeauthoryear{Xu, Ba, Kiros, Cho, Courville, Salakhudinov, Zemel, and Bengio}{Xu et~al\mbox{.}}{2015}]%
        {xu2015show}
\bibfield{author}{\bibinfo{person}{Kelvin Xu}, \bibinfo{person}{Jimmy Ba}, \bibinfo{person}{Ryan Kiros}, \bibinfo{person}{Kyunghyun Cho}, \bibinfo{person}{Aaron Courville}, \bibinfo{person}{Ruslan Salakhudinov}, \bibinfo{person}{Rich Zemel}, {and} \bibinfo{person}{Yoshua Bengio}.} \bibinfo{year}{2015}\natexlab{}.
\newblock \showarticletitle{Show, attend and tell: Neural image caption generation with visual attention}. In \bibinfo{booktitle}{\emph{International conference on machine learning}}. PMLR, \bibinfo{pages}{2048--2057}.
\newblock


\bibitem[\protect\citeauthoryear{Yan, Liu, Luo, Qiu, Wang, and Tang}{Yan et~al\mbox{.}}{2017}]%
        {yan2017unconstrained}
\bibfield{author}{\bibinfo{person}{Sijie Yan}, \bibinfo{person}{Ziwei Liu}, \bibinfo{person}{Ping Luo}, \bibinfo{person}{Shi Qiu}, \bibinfo{person}{Xiaogang Wang}, {and} \bibinfo{person}{Xiaoou Tang}.} \bibinfo{year}{2017}\natexlab{}.
\newblock \showarticletitle{Unconstrained fashion landmark detection via hierarchical recurrent transformer networks}. In \bibinfo{booktitle}{\emph{Proceedings of the 25th ACM international conference on Multimedia}}. \bibinfo{pages}{172--180}.
\newblock


\bibitem[\protect\citeauthoryear{Yang, Dai, Yang, Carbonell, Salakhutdinov, and Le}{Yang et~al\mbox{.}}{2019}]%
        {yang2019xlnet}
\bibfield{author}{\bibinfo{person}{Zhilin Yang}, \bibinfo{person}{Zihang Dai}, \bibinfo{person}{Yiming Yang}, \bibinfo{person}{Jaime Carbonell}, \bibinfo{person}{Russ~R Salakhutdinov}, {and} \bibinfo{person}{Quoc~V Le}.} \bibinfo{year}{2019}\natexlab{}.
\newblock \showarticletitle{Xlnet: Generalized autoregressive pretraining for language understanding}.
\newblock \bibinfo{journal}{\emph{Advances in neural information processing systems}}  \bibinfo{volume}{32} (\bibinfo{year}{2019}).
\newblock


\bibitem[\protect\citeauthoryear{Yuan and Lam}{Yuan and Lam}{2021a}]%
        {yuan2021conversational}
\bibfield{author}{\bibinfo{person}{Yifei Yuan} {and} \bibinfo{person}{Wai Lam}.} \bibinfo{year}{2021}\natexlab{a}.
\newblock \showarticletitle{Conversational Fashion Image Retrieval via Multiturn Natural Language Feedback}. In \bibinfo{booktitle}{\emph{Proceedings of the 44th International ACM SIGIR Conference on Research and Development in Information Retrieval}}.
\newblock


\bibitem[\protect\citeauthoryear{Yuan and Lam}{Yuan and Lam}{2021b}]%
        {Yuan2021SentimentAO}
\bibfield{author}{\bibinfo{person}{Yifei Yuan} {and} \bibinfo{person}{Wai Lam}.} \bibinfo{year}{2021}\natexlab{b}.
\newblock \showarticletitle{Sentiment Analysis of Fashion Related Posts in Social Media}.
\newblock \bibinfo{journal}{\emph{Proceedings of the Fifteenth ACM International Conference on Web Search and Data Mining}} (\bibinfo{year}{2021}).
\newblock
\urldef\tempurl%
\url{https://api.semanticscholar.org/CorpusID:244117093}
\showURL{%
\tempurl}


\bibitem[\protect\citeauthoryear{Zhang, Zhang, Li, and Qiao}{Zhang et~al\mbox{.}}{2016}]%
        {zhang2016joint}
\bibfield{author}{\bibinfo{person}{Kaipeng Zhang}, \bibinfo{person}{Zhanpeng Zhang}, \bibinfo{person}{Zhifeng Li}, {and} \bibinfo{person}{Yu Qiao}.} \bibinfo{year}{2016}\natexlab{}.
\newblock \showarticletitle{Joint face detection and alignment using multitask cascaded convolutional networks}.
\newblock \bibinfo{journal}{\emph{IEEE Signal Processing Letters}} \bibinfo{volume}{23}, \bibinfo{number}{10} (\bibinfo{year}{2016}), \bibinfo{pages}{1499--1503}.
\newblock


\bibitem[\protect\citeauthoryear{Zhang, Deng, Li, Yuan, Bing, and Lam}{Zhang et~al\mbox{.}}{2021a}]%
        {emnlp21-absa}
\bibfield{author}{\bibinfo{person}{Wenxuan Zhang}, \bibinfo{person}{Yang Deng}, \bibinfo{person}{Xin Li}, \bibinfo{person}{Yifei Yuan}, \bibinfo{person}{Lidong Bing}, {and} \bibinfo{person}{Wai Lam}.} \bibinfo{year}{2021}\natexlab{a}.
\newblock \showarticletitle{Aspect Sentiment Quad Prediction as Paraphrase Generation}. In \bibinfo{booktitle}{\emph{{EMNLP} 2021}}. \bibinfo{pages}{9209--9219}.
\newblock


\bibitem[\protect\citeauthoryear{Zhang, Li, Deng, Bing, and Lam}{Zhang et~al\mbox{.}}{2021b}]%
        {acl21-absa}
\bibfield{author}{\bibinfo{person}{Wenxuan Zhang}, \bibinfo{person}{Xin Li}, \bibinfo{person}{Yang Deng}, \bibinfo{person}{Lidong Bing}, {and} \bibinfo{person}{Wai Lam}.} \bibinfo{year}{2021}\natexlab{b}.
\newblock \showarticletitle{Towards Generative Aspect-Based Sentiment Analysis}. In \bibinfo{booktitle}{\emph{{ACL/IJCNLP} 2021}}. \bibinfo{pages}{504--510}.
\newblock


\bibitem[\protect\citeauthoryear{Zhao, Deng, Yang, Wang, Zhang, Cheng, Lam, Shen, and Xu}{Zhao et~al\mbox{.}}{2023}]%
        {re-survey}
\bibfield{author}{\bibinfo{person}{Xiaoyan Zhao}, \bibinfo{person}{Yang Deng}, \bibinfo{person}{Min Yang}, \bibinfo{person}{Lingzhi Wang}, \bibinfo{person}{Rui Zhang}, \bibinfo{person}{Hong Cheng}, \bibinfo{person}{Wai Lam}, \bibinfo{person}{Ying Shen}, {and} \bibinfo{person}{Ruifeng Xu}.} \bibinfo{year}{2023}\natexlab{}.
\newblock \showarticletitle{A Comprehensive Survey on Deep Learning for Relation Extraction: Recent Advances and New Frontiers}.
\newblock \bibinfo{journal}{\emph{CoRR}}  \bibinfo{volume}{abs/2306.02051} (\bibinfo{year}{2023}).
\newblock


\end{thebibliography}


\end{document}